%% file: arxiv.tex
\documentclass[sigconf, nonacm, pdfa]{acmart}

\usepackage[utf8]{inputenc}
\usepackage{amsmath,amsfonts}
\usepackage{amsthm}
\usepackage{textcomp}
\usepackage{balance}
\usepackage{color,xcolor}
\usepackage{multirow}
\usepackage{array}
\usepackage{enumitem}
\usepackage{epstopdf}
\usepackage{booktabs}
\usepackage{threeparttable}
\usepackage{bm}
\usepackage{graphicx}
\usepackage{subfigure}
\usepackage{caption}
\usepackage{gensymb}

\newtheoremstyle{mydefn}
{}{}
{\it}       
{0pt}       
{\bfseries} 
{:~}        
{0pt}       
{}          
\theoremstyle{mydefn}

\newtheorem{definition}{Definition}[section]

\newtheorem{lemma}{Lemma}[section]

\newtheorem{corollary}{Corollary}[section]

\newtheoremstyle{myexample}
{}{}
{}          
{0pt}       
{\bfseries} 
{:~}        
{0pt}       
{}          
\theoremstyle{myexample}

\newtheorem{example}{Example}[section]

\DeclareMathOperator*{\argmin}{arg\,min}
\DeclareMathOperator*{\argmax}{arg\,max}

\renewenvironment{proof}{\smallskip\noindent{\bfseries Proof:}}{\qed} 
\newenvironment{sproof}{\smallskip\noindent{\bfseries Proof Sketch:}}{\qed}

\renewcommand{\paragraph}[1]{\vspace{0.25em}\noindent\textbf{#1.}}
\renewcommand{\subparagraph}[1]{\vspace{0.25em}\noindent\textit{\underline{#1.}}}

\newcommand{\keypoint}[1]{\textit{\underline{#1:}}}
\newcommand{\revise}[1]{{\color{black} #1}}

\newcommand{\norm}[1]{\Vert #1 \Vert}
\newcommand{\num}[1]{\vert #1 \vert}

\newcommand\vldbdoi{XX.XX/XXX.XX}
\newcommand\vldbpages{XXX-XXX}
\newcommand\vldbvolume{XX}
\newcommand\vldbissue{X}
\newcommand\vldbyear{20XX}
\newcommand\vldbauthors{\authors}
\newcommand\vldbtitle{\shorttitle} 
\newcommand\vldbavailabilityurl{https://anonymous.4open.science/r/CTS-2376/}
\newcommand\vldbpagestyle{plain} 

\begin{document}

\input{01_title}

\input{02_abstract}
\pagestyle{\vldbpagestyle}
\begingroup\small\noindent\raggedright\textbf{ACM Reference Format:}\\
\vldbauthors. \vldbtitle. ACM Conference, \vldbvolume(\vldbissue): \vldbpages, \vldbyear.\\
\href{https://doi.org/\vldbdoi}{doi:\vldbdoi}
\endgroup
\begingroup
\renewcommand\thefootnote{}\footnote{\noindent
Permission to make digital or hard copies of all or part of this work for personal or classroom use is granted without fee provided that copies are not made or distributed for profit or commercial advantage and that copies bear this notice and the full citation on the first page. Copyrights for components of this work owned by others than ACM must be honored. Abstracting with credit is permitted. To copy otherwise, or republish, to post on servers or to redistribute to lists, requires prior specific permission and/or a fee. Request permissions from permissions@acm.org. \\ 
\raggedright ACM Conference, Vol. \vldbvolume, No. \vldbissue\ %
ISSN XXXX-XXXX. \\
\href{https://doi.org/\vldbdoi}{doi:\vldbdoi} \\
}\addtocounter{footnote}{-1}\endgroup


\input{03_introduction}
\input{04_related_work}
\input{05_problem}
\input{06_framework}
\input{07_evaluation}
\input{08_experiments}

\input{09_conclusions}

\balance
\bibliographystyle{ACM-Reference-Format}
\bibliography{arxiv}

\end{document}

%% file: 01_title.tex
\title{Towards Controllable Time Series Generation}

\author{Yifan Bao}
\orcid{0009-0000-9672-0747}
\affiliation{
 \institution{National University of Singapore}
 \country{}
}
\email{yifan\_bao@comp.nus.edu.sg}

\author{Yihao Ang}
\orcid{0009-0009-1564-4937}
\affiliation{
 \institution{National University of Singapore}
 \institution{NUS Research Institute in Chongqing}
 \country{}
}
\email{yihao\_ang@comp.nus.edu.sg}

\author{Qiang Huang}
\orcid{0000-0003-1120-4685}
\affiliation{
 \institution{National University of Singapore}
 \country{}
}
\email{huangq@comp.nus.edu.sg}

\author{Anthony K. H. Tung}
\orcid{0000-0001-7300-6196}
\affiliation{
 \institution{National University of Singapore}
 \country{}
}
\email{atung@comp.nus.edu.sg}

\author{Zhiyong Huang}
\orcid{0000-0002-1931-7775}
\affiliation{
 \institution{National University of Singapore}
 \institution{NUS Research Institute in Chongqing}
 \country{}
}
\email{huangzy@comp.nus.edu.sg}

%% file: 02_abstract.tex
\begin{abstract}
Time Series Generation (TSG) has emerged as a pivotal technique in synthesizing data that accurately mirrors real-world time series, becoming indispensable in numerous applications. 
Despite significant advancements in TSG, its efficacy frequently hinges on having large training datasets. This dependency presents a substantial challenge in data-scarce scenarios, especially when dealing with rare or unique conditions.
To confront these challenges, we explore a new problem of Controllable Time Series Generation (CTSG), aiming to produce synthetic time series that can adapt to various external conditions, thereby tackling the data scarcity issue.

In this paper, we propose \textbf{C}ontrollable \textbf{T}ime \textbf{S}eries (\textsf{CTS}), an innovative VAE-agnostic framework tailored for CTSG. 
A key feature of \textsf{CTS} is that it decouples the mapping process from standard VAE training, enabling precise learning of a complex interplay between latent features and external conditions.
Moreover, we develop a comprehensive evaluation scheme for CTSG.
Extensive experiments across three real-world time series datasets showcase \textsf{CTS}'s exceptional capabilities in generating high-quality, controllable outputs. This underscores its adeptness in seamlessly integrating latent features with external conditions. 
Extending \textsf{CTS} to the image domain highlights its remarkable potential for explainability and further reinforces its versatility across different modalities.
\end{abstract}

\maketitle

%% file: 03_introduction.tex
\section{Introduction}
\label{sec:intro}

In the evolving area of time series, Time Series Generation (TSG) has become pivotal for synthesizing data that closely emulates real-world time series, preserving essential temporal patterns and multidimensional correlations.  
TSG plays a vital role in diverse applications including classification \cite{li2022ips, ding2022towards}, anomaly detection \cite{cad, campos2021unsupervised}, domain transfer \cite{sasa, tsgbench}, and privacy protection \cite{pategan}.
Within the spectrum of TSG methods, Generative Adversarial Networks (GANs) \cite{rcgan, timegan, rtsgan, cosci-gan, aec-gan} excel in capturing intricate time series characteristics. 
Parallel to this, Variational AutoEncoders (VAEs) \cite{timevae, timevqvae} have become a leading technique in the field.
Their primary strength lies in adeptly balancing high data fidelity and statistical consistency in latent spaces, greatly enhancing the interpretability of the generated data.
A recent TSG benchmark study \cite{tsgbench} substantiates the superiority of VAE-based methods, demonstrating their exceptional performance across multiple datasets and tasks.

While TSG demonstrates proficiency in generating time series akin to real-world data, its efficacy is often contingent on the availability of large datasets for training \cite{rtsgan, DoppelGANger, tsgbench}.
This requirement poses a significant challenge in scarce data scenarios, and it becomes particularly pronounced when data is generated under rare, unique, or hard-to-replicate conditions, leading to gaps in comprehensive data analysis and understanding.
For example, in environmental science, modeling the impact of diverse outdoor conditions using air pollutant time series is crucial for developing effective ecological and public health policies \cite{zheng2015forecasting}. 
However, the rarity of extreme weather events leads to a scarcity of representative data, impeding in-depth analysis during critical periods.
In the field of manufacturing, monitoring machinery through sensory time series is key to optimizing maintenance schedules. 
Yet, as machines predominantly operate under standard conditions, the paucity of data on anomalous states renders timely response difficult \cite{cad}.

Contemporary methods for TSG encounter significant challenges when dealing with scarce data, primarily due to two inherent limitations:
(1) Realistic Generation from Scarce Data: Many existing TSG methods, originally designed for data-rich environments, struggle to capture the full intricacies of datasets when data is sparse, hindering their capability to accurately reproduce underlying patterns.
(2) Controllability: For TSG applications to be effective, producing data that adheres to specific, occasionally rare, user-defined conditions is crucial.
This becomes particularly hard when such conditions are underrepresented in the original dataset.

In this paper, we investigate a new problem, Controllable Time Series Generation (CTSG), addressing the challenge of data scarcity.
CTSG focuses on training a generative model using a time series dataset with various external conditions. 
This model is designed to learn patterns in historical data and adjust to user-specified changes in external conditions. 
As a result, it generates synthetic time series that not only reflect these altered conditions but also maintain a balance of realism and contextual relevance.
According to different use cases, CTSG can be broadly categorized into two scenarios based on external conditions:
(1) Interpolation: CTSG generates time series under conditions within the known range, enhancing data granularity and aiding in more refined learning for downstream tasks.
(2) Extrapolation: In this scenario, CTSG extends beyond the existing data range to accommodate user-specified conditions. It is capable of generating time series without direct historical equivalents while ensuring relevance and consistency within the domain's context.
Drawing from the examples above, environmental scientists can utilize CTSG to simulate data under extreme weather conditions, paving the way for more proactive and informed response strategies.
Industrial engineers can generate machine sensory data based on anomalous states, enabling early anomaly detection and mitigation of potential failures.

Some earlier efforts \cite{cgan, rcgan} have focused on Conditional TSG, primarily using conditional generative models. They incorporate class labels into model training to learn joint distributions, facilitating the generation of synthetic time series across different classes.
However, their reliance on comprehensive training data limits their effectiveness in scenarios with scarce data. The specific, user-defined conditions may be sparse or absent in the training dataset, rendering them less suitable for CTSG.
The computer vision field has increasingly embraced disentanglement learning to tackle the controllability issue, frequently applying it in conjunction with VAEs \cite{locatello2019challenging, trauble2021disentangled}.  
Advanced supervised disentangled VAEs \cite{maaloe2016auxiliary, siddharth2017learning, li2019disentangled, khemakhem2020variational, ding2020guided, mita2021identifiable, joy2021capturing, zou2022joint} have been developed to ensure that latent features are not only independent but also meaningfully correlated with external conditions. 
They enable precise manipulation of latent features by integrating these conditions during VAE training, thus enhancing controllability.
Nonetheless, adapting these methods to CTSG presents significant challenges:
\begin{enumerate}[nolistsep,leftmargin=25pt]
  \item \keypoint{Tradeoff in Generation Quality}
  These VAEs jointly optimize generation quality and latent feature independence, leading to a trade-off in generation fidelity and reduced details, impeding users' understanding of generated data \cite{shao2022rethinking}.

  \item \keypoint{Intricate Condition Relationships}
  They typically strive for independent mapping between latent features and external conditions. 
  Nevertheless, these conditions often exhibit inter-correlations, i.e., changing one condition might unintentionally influence others \cite{li2019learning, trauble2021disentangled}, complicating capturing accurate relationships among external conditions.
  
  \item \keypoint{Data Modality Limitation}
  Disentangled VAEs, primarily developed for image data, are optimized to manage the distinct attributes of visual information.
  Yet, time series data present different characteristics and complexities, such as temporal dependencies and variable frequencies \cite{tonekaboni2020went, leung2022temporal}, posing a significant challenge to adapting these models to CTSG.

  \item \keypoint{Explainability Gap}
  The complexity of these VAEs, especially those using end-to-end deep neural networks, tends to obscure transparency and interpretability, which are vital for understanding and trusting the model’s outputs.
\end{enumerate}

\paragraph{Our Contributions}
To address the above challenges, we introduce \textbf{C}ontrollable \textbf{T}ime \textbf{S}eries (\textsf{CTS}), a novel VAE-agnostic framework for CTSG. To the best of our knowledge, \textsf{CTS} is the first foray into this burgeoning field.
It innovatively decouples the mapping process from standard VAE training. This separation enables customized learning of the intricate interactions between latent features and external conditions, thereby enhancing the controllability and finesse of the generation process.
The \textsf{CTS} framework is characterized by the following three core properties:
\vspace{0.2em}
\begin{enumerate}[nolistsep,leftmargin=25pt]
  \item \keypoint{VAE-Agnostic Design} 
  The distinct separation of the mapping function from VAE training in \textsf{CTS} imparts remarkable flexibility to accommodate different VAE variants. This adaptability enables \textsf{CTS} to harness a range of cutting-edge VAE-based models for TSG, thereby guaranteeing optimal generation fidelity.

  \item \keypoint{Mapping Function Integration} 
  \textsf{CTS} incorporates a mapping function that capitalizes on the VAE's latent features to discern intricate relationships with external conditions. Crucially, it does not mandate complete independence of latent features, enabling high controllability and a richer interpretation of how various external conditions interplay.

  \item \keypoint{Versatility and Explainability} 
  \textsf{CTS}'s VAE-agnostic design enables versatility across various modalities, surpassing typical modality-specific limitations.
  Moreover, its data selection module, identifying similar data under varied conditions, combined with a white-box regression model for condition mapping, significantly boosts explainability.
\end{enumerate}
\vspace{0.2em}

Evaluating CTSG presents challenges, primarily due to (1) existing metrics being tailored for traditional TSG, and (2) the focus on one-to-one mappings in disentangled VAEs \cite{deng2020disentangled, chen2018isolating, kim2018disentangling}, which may overlook complex interactions in multidimensional conditions.
To remedy these issues, we design a new evaluation scheme for CTSG, applicable to both interpolation and extrapolation:
(1) \emph{Generation Fidelity}: we evaluate how closely the generated series mirrors the original, using established TSG metrics to measure the fidelity.
(2) \emph{Attribute Coherence}: we assess the preservation of essential inherent attributes in the generated series, ensuring that these attributes remain coherent and consistent.
(3) \emph{Controllability}: we examine how well the generated series aligns with user-modified conditions, especially applicable in scenarios without ground truth.

We assess \textsf{CTS} on three real-world datasets, examining its capability in interpolation through classification and extrapolation in external conditions via anomaly detection. 
Our extensive results, including ablation studies and sensitivity tests, validate \textsf{CTS}'s exceptional generation quality and controllability, confirming its expertise in bridging latent features with external conditions.
Furthermore, we extend \textsf{CTS} to the image domain, testing it on three large-scale datasets and comparing it against three advanced supervised disentangled VAEs.
This extension showcases \textsf{CTS}'s versatility across different domains and its superior explainability.

\paragraph{Organizations}
The rest of this paper is organized as follows. 
Section \ref{sec:related} reviews the related work. 
The problem of CTSG is formulated in Section \ref{sec:problem}. 
We introduce \textsf{CTS} in Section \ref{sec:framework} and develop a CTSG evaluation scheme in Section \ref{sec:eval}. 
Experimental results are presented and analyzed in Section \ref{sec:expt}.
We conclude this work in Section \ref{sec:conclusion}.

%% file: 04_related_work.tex
\section{Related Work}
\label{sec:related}

\noindent\textbf{Time Series Generation (TSG).}
In the field of TSG, early approaches integrated standard GAN frameworks from the domain of image generation with neural networks like RNN and LSTM \cite{crnngan, rcgan, t-cgan, timegan, rtsgan}.
Recent advancements introduce novel metrics or loss functions to mirror particular temporal patterns accurately \cite{sig-gan, c-sig-gan}. 
Furthermore, some methods improve upon classic GANs by incorporating extra components, such as additional discriminators, classification layers, error correction mechanisms, and data augmentation techniques, to produce time series data with precise temporal characteristics \cite{aec-gan, cosci-gan, psa-gan}.
While GAN-based approaches are proficient in TSG, they often pose training challenges and require substantial resources and time \cite{gtgan}.

Compared to the prevalence of GAN-based studies, there are relatively fewer methods employing VAEs. 
However, these approaches effectively utilize variational inference to capture the intricate temporal dynamics characteristic of time series data \cite{timevae, crvae, timevqvae}.
For example, \textsf{TimeVAE} \cite{timevae} employs convolutional techniques to balance generation quality and operational efficiency.
Recently, \textsf{TimeVQVAE} \cite{timevqvae} uses the Short-Time Fourier Transform (STFT) to transform time series into their frequency components. Despite its higher computational demands, which can limit scalability for larger datasets \cite{tsgbench}, its integration of Vector Quantization (VQ) with VAEs \cite{vqvae} significantly enhances the fidelity of generated data. 

GANs and VAEs, while powerful, do not inherently model the probability density function of time series data due to the extensive computational requirements of encompassing all potential latent representations.
To mitigate this limitation, mixed-type approaches are gaining traction in TSG. 
These methods often combine explicit likelihood models or Ordinary Differential Equations (ODEs) \cite{NODE, ode-rnn, gtgan, fourier}, employing architectures with coupling layers that facilitate a calculable Jacobian determinant and ensure reversibility.

A recent benchmark study, \textsf{TSGBench} \cite{tsgbench}, reveals the efficacy of VAE-based methods in TSG, showcasing their outstanding performance across a wide range of datasets and evaluation metrics.
This evidence emphasizes the potential of VAEs not just to mimic but even surpass the capabilities of real-world data in various scenarios.
Motivated by these insights, we introduce the \textsf{CTS} framework, utilizing the strength of VAEs to advance the field of CTSG.

\vspace{0.1em}
\paragraph{Controllable Image Generation by VAEs}
As VAEs have demonstrated their prowess in image generation, 
prior work in the image domain often employs unsupervised learning to achieve distinct representations \cite{higgins2017beta, burgess2018understanding, kim2018disentangling, chen2018isolating, kumar2017variational, rubenstein2018learning, zhao2019infovae, locatello2019fairness, locatello2019disentangling}. 
Yet, \citet{locatello2019challenging} highlighted that supervision often enhances disentanglement. 
As a result, numerous approaches have incorporated explicit supervision to improve the disentanglement of the latent representations \cite{reed2014learning, zhu2014multi, yang2015weakly, kulkarni2015deep, cheung2014discovering, mathieu2016disentangling, siddharth2017learning, suter2018interventional, kingma2014semi}.
Since the labeled data is often inadequate, many recent methods also explore weakly-supervised strategies to address this issue based on a limited number of labels or group structures \cite{goyal2019recurrent, schmidt2007learning, bengio2019meta, ke2019learning, klindt2020towards, shu2020weakly, locatello2020weakly, bouchacourt2018multi, nemeth2020adversarial}.
In addition, practical external conditions often exhibit interdependencies \cite{trauble2021disentangled}. 
As such, several pioneering methods attempt to bridge the gap between disentangled latent features and correlated conditions \cite{li2019learning, trauble2021disentangled}.

Recent approaches have been geared towards providing intuitive and interpretable control by associating external image conditions with disentangled latent features \cite{joy2021capturing, mita2021identifiable, ding2020guided, zou2022joint}. 
For instance, \textsf{CCVAE} \cite{joy2021capturing} captures label characteristics in the latent space rather than directly associating them with label values.
\textsf{Guided-VAE} \cite{ding2020guided} signals the latent encoding, offering supervised and unsupervised disentanglement.
\textsf{IDVAE} \cite{mita2021identifiable} learns an optimal
conditional prior, ensuring better control via enhanced regularization of latent features. 
While these methods connect latent features to external conditions as supervision, \textsf{CTS} stands out as the first framework to separate the mapping function from the conventional VAE training. 
This unique design allows \textsf{CTS} to develop more complex and nuanced associations between latent features and external conditions.

In the realm of time series, the concept of disentanglement has been explored but to a lesser extent. Current research predominantly concentrates on areas such as domain adaptation and clustering \cite{som-vae, li2022towards}.
To the best of our knowledge, no current work explicitly explores the CTSG process with external conditions.

%% file: 05_problem.tex
\section{Problem Formulation}
\label{sec:problem}

\input{tables/notations}

Consider $\bm{X}$ as a training time series dataset gathered from a specific data source. Let $\bm{x}$ represent an individual time series contained in $\bm{X}$. 
Table \ref{tab:notations} summarizes the frequently used notations throughout this paper.
In the following, we define the problem of Time Series Generation (TSG) using VAEs.
\vspace{-0.4em}
\begin{definition}[Time Series Generation (TSG) using VAEs]
\label{def:tsg-by-vae}
  Consider a training time series dataset $\bm{X}$ and a VAE model, $\bm{g}$, composed of an encoder \textbf{enc}$(\cdot)$ and a decoder \textbf{dec}$(\cdot)$. 
  For any time series $\bm{x} \in \bm{X}$, the objective of TSG using VAEs is to minimize the difference between $\bm{x}$ and its reconstructed version $\bm{x^\prime}$, derived from $\bm{g}$. 
  Specifically, $\bm{x^\prime} =$ \textbf{dec}$(\bm{z}) =$ \textbf{dec}$(\bm{\mu} + \bm{\sigma} \odot \bm{\epsilon})$, where $\bm{\mu}$ and $\bm{\sigma}$ are the mean latent vector and the log variance vector, respectively, produced by \textbf{enc}$(\bm{x})$. 
  The symbol $\odot$ indicates element-wise multiplication. 
  The latent vector of $\bm{x}$, represented by $\bm{z}$, is determined by \textbf{enc}$(\cdot)$ through the reparameterization trick. 
  Additionally, $\bm{\epsilon} \sim \mathcal{N}(\bm{0}, \bm{I})$ serves as a random Gaussian noise vector for the sampling purpose.
\end{definition}

\vspace{-0.35em}
In numerous scenarios, ranging from air pollutant simulation to machine sensory data generation, their time series $\bm{x}$ is typically associated with an external condition vector $\bm{c} = (c_1, \cdots, c_m)$. Here, $c_i$ ($1 \leq i \leq m$) denotes the $i$-th condition value of $\bm{x}$. 
To enable Controllable Time Series Generation (CTSG), we propose a mapping function $\bm{f}$ to delineate the intricate relationships between $\bm{c}$ and the latent vector $\bm{z}$ derived from a VAE model $\bm{g}$. Formally,
\begin{definition}[Controllable Time Series Generation (CTSG)]
\label{def:ctsg-by-vae-ec}
  Consider a training time series dataset $\bm{X}$ and its corresponding set $\bm{C}$ of external condition vectors. Let $\bm{g}$ be a VAE model trained on $\bm{X}$. 
  For each time series $\bm{x} \in \bm{X}$ with its associated external condition vector $\bm{c}$, the objective of CTSG is to establish a mapping function, $\bm{f}(\cdot)$, that bridges $\bm{c}$ and the latent vector $\bm{z}$ derived from the encoder of $\bm{g}$, i.e., $\bm{z} = \bm{f}(\bm{c})$. 
  Consequently, for any user-input time series $\bm{x_0}$ and its external condition vector $\bm{c_0}$, the decoder should generate a new time series, $\bm{x_0^\prime}$, that aligns with a user-specified external condition vector $\bm{c_0^\prime}$, i.e., $\bm{x_0^\prime} = \bm{g}(\bm{z_0^\prime}; \bm{x_0}) = \bm{g}(\bm{f}(\bm{c_0^\prime}); \bm{x_0})$.
\end{definition}

%% file: tables/notations.tex
\begin{table}[t]
\centering
\footnotesize
\captionsetup{skip=0.5em}
\caption{List of frequently used notations.}
\label{tab:notations}
\begin{tabular}{p{0.14\columnwidth}p{0.78\columnwidth}} \toprule
  \textbf{Symbol} & \textbf{Description} \\ \midrule
  $\bm{X}$, $d_r$ & A set of $n$ training time series data, each with $d_r$ raw features \\
  $\bm{C}$, $m$ & A set of $n$ condition vectors, each with $m$ external conditions \\
  $\bm{Z}$, $d_l$ & A set of $n$ latent vectors, each with $d_l$ latent features \\
  $\bm{x}$, $\bm{c}$, $\bm{z}$ & A training time series, its external condition and latent vectors \\ 
  $\bm{x_0}$, $\bm{c_0}$, $\bm{z_0}$ & A user-input time series, its external condition and latent vectors \\
  $\bm{g}$ & A VAE model for Time Series Generation \\
  $\bm{c_0^\prime}$ & The user-specified external condition vector \\
  $\bm{x_0^\prime}$, $\bm{z_0^\prime}$ & The newly generated time series and its latent vector \\
  $\bm{\mu}$, $\bm{\sigma}$, $\bm{\epsilon}$ & The mean latent vector, log variance vector, and random Gaussian noise vector \\ 
  $\bm{q_\phi}(\bm{z}|\bm{x})$ & A VAE encoder, where $\bm{\phi}$ is a vector of VAE encoder parameters \\ 
  $\bm{p_\theta}(\bm{x}|\bm{z})$ & A VAE decoder, where $\bm{\theta}$ is a vector of VAE decoder parameters \\ 
  $\bm{f}$ & The mapping function between the external condition vector $\bm{c}$ and the mean latent vector $\bm{\mu}$ \\
  $k$, $k_1$, $k_2$ & The number of clusters, the number of selected clusters in \textsf{DCS}, and the number of most similar time series in each cluster in \textsf{NNS} \\ 
  $\bm{X_s}$, $\bm{C_s}$, $\bm{Z_s}$ & A time series subset from Data Selection and their external condition vectors and latent vectors, where $\num{\bm{X_s}} = k_1 \cdot k_2$ \\ 
  $\norm{\bm{\mu}-\bm{\mu^\prime}}_2^2$ & The $l_2$ loss between the true mean latent vector $\bm{\mu}$ and the predicted mean latent vector $\bm{\mu^\prime}$ \\
  \bottomrule 
\end{tabular}
\end{table}
\setlength{\textfloatsep}{1.0em}

%% file: 06_framework.tex
\section{The \textsf{CTS} Framework}
\label{sec:framework} 

\subsection{Overview}
\label{sec:framework:overview}

\begin{figure}[h]
\centering
\vspace{-0.25em}
\includegraphics[width=0.99\columnwidth]{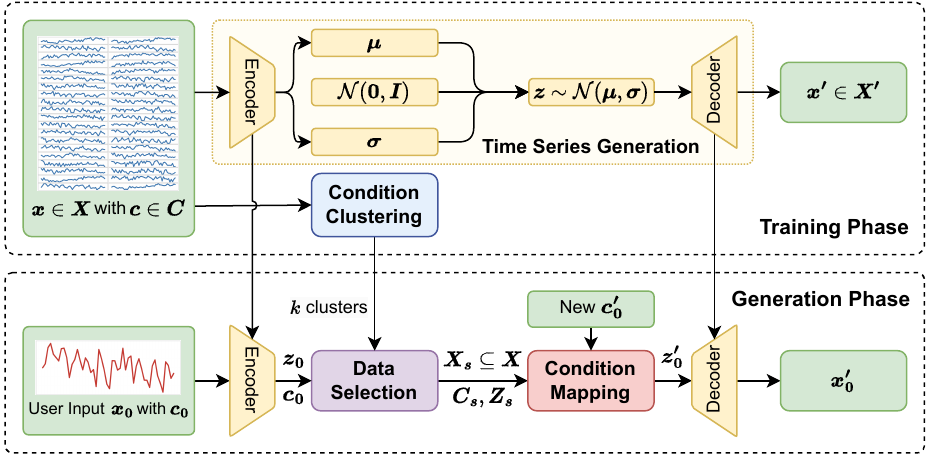}
\vspace{-1.0em}
\caption{The overall pipeline of the \textsf{CTS} framework.}
\label{fig:framework}
\vspace{-0.75em}
\end{figure}

In this section, we present \textsf{CTS}, a novel VAE-agnostic framework developed for CTSG. 
Unlike the advanced supervised disentangled VAEs in Controllable Image Generation \cite{li2019disentangled, khemakhem2020variational, ding2020guided, mita2021identifiable, joy2021capturing, zou2022joint}, the core concept of \textsf{CTS} revolves around decoupling the mapping process from VAE training.
This strategy allows for the development of the mapping function in a white-box manner, enhancing \textsf{CTS}'s explainability and adaptability to cater to diverse user-specific needs.
The overall pipeline of \textsf{CTS} is depicted in Figure~\ref{fig:framework}.

\textsf{CTS} operates in two main phases: training and generation.
In the training phase, the \emph{Time Series Generation} module employs a set of $n$ training time series, denoted as $\bm{X}$, to learn a leading VAE model to produce reconstructed time series $\bm{x^\prime}$ for any input $\bm{x}$. 
Subsequently, the \emph{Condition Clustering} module partitions $\bm{X}$ into $k$ clusters based on their associated external condition vectors $\bm{C}$ to facilitate controllability.
The generation phase begins with a user-input time series $\bm{x_0}$ and its external condition vector $\bm{c_0}$.
The \emph{Data Selection} module identifies a subset $\bm{X_s}$ from $\bm{X}$ that highlights a variety of external conditions relative to $\bm{c_0}$ and exhibits visual resemblance to $\bm{x_0}$. 
Finally, the \emph{Condition Mapping} module explicitly learns a mapping function $\bm{f}$ between external conditions and latent features. 
This feature enables users to modify and interpret a chosen external condition $\bm{c_0^\prime}$, facilitating the generation of new time series $\bm{x_0^\prime}$ throughout the generative process.
Next, we shall provide a detailed exposition of the training and generation phases.

\subsection{Training Phase}
\label{sec:framework:train}

\noindent\textbf{Time Series Generation.}
Given a set $\bm{X}$ of $n$ training time series, this module aims to learn a VAE model for TSG. 
We assume that each time series $\bm{x} \in \bm{X}$ is sampled from a generative process $\bm{p}(\bm{x}|\bm{z})$, where $\bm{z}$ refers to the latent vector in a latent feature space. 
In practice, $\bm{z}$ and the generative process $\bm{p}(\bm{x}|\bm{z})$ are often unknown in advance. 
Thus, we aim to simultaneously train an encoder $\bm{q_\phi}(\bm{z}|\bm{x})$ and a decoder $\bm{p_\theta}(\bm{x}|\bm{z})$, where $\bm{\phi}$ and $\bm{\theta}$ are two vectors as the model parameters. 
Here, the encoder is a neural network that outputs the parameters for Gaussian distribution $\bm{q_\phi}(\bm{z}|\bm{x}) = \mathcal{N}(\bm{\mu},\bm{\sigma})$; 
the decoder is a deterministic neural network $\bm{\eta_\theta}(\bm{z})$, where the generative density $\bm{p_\theta}(\bm{x}|\bm{z})$ can be supposed to be subject to a Gaussian distribution, i.e., $\bm{p_\theta}(\bm{x}|\bm{z}) = \mathcal{N}(\bm{\eta_\theta}(\bm{z}), \bm{\Sigma}^2 \bm{I})$. 
We use the evidence lower bound (ELBO) below to train the encoder and decoder: 
\begin{displaymath}
\mathcal{L}(\bm{\theta},\bm{\phi}; \bm{x}) = \mathbb{E}_{\bm{q_\phi}(\bm{z}|\bm{x})}[\log \bm{p_\theta}(\bm{x}|\bm{z})] - D_{KL}(\bm{q_\phi}(\bm{z}|\bm{x}) \Vert \bm{p}(\bm{z})),
\end{displaymath}
where $\bm{p}(\bm{z})$ is the prior, which is assumed to be $\mathcal{N}(\bm{0},\bm{I})$. 
The first term $\mathbb{E}_{\bm{q_\phi}(\bm{z}|\bm{x})}[\log \bm{p_\theta}(\bm{x}|\bm{z})]$ can be reduced to the standard reconstruction loss $\mathbb{E}_{\bm{q_\phi}(\bm{z}|\bm{x})}[\norm{\bm{x} - \bm{\eta_\theta}(\bm{z})}_2^2]$ (up to an additive constant). 
The second term $D_{KL}(\bm{q_\phi}(\bm{z}|\bm{x}) \Vert \bm{p}(\bm{z}))$ is the Kullback-Leibler (KL) divergence \cite{csiszar1975divergence} between $\bm{q_\phi}(\bm{z}|\bm{x})$ and $\bm{p}(\bm{z})$.
The encoder and decoder can be \emph{jointly} optimized over the training time series $\bm{x} \in \bm{X}$ by maximizing the ELBO, i.e., 
\begin{displaymath}
\bm{\theta^*}, \bm{\phi^*} = {\argmax}_{\bm{\theta},\bm{\phi}} \mathcal{L}(\bm{\theta}, \bm{\phi}; \bm{x}).
\end{displaymath}

This module employs the encoder $\bm{q_{\phi}}(\bm{z}|\bm{x})$ to generate the latent vector $\bm{z_0}$ for the user-input time series $\bm{x_0}$, aiding the Data Selection process.
Concurrently, it utilizes the decoder $\bm{p_{\theta}}(\bm{x}|\bm{z})$ to produce the reconstructed time series $\bm{x_0^\prime}$ using the latent vector $\bm{z_0^\prime}$ derived from Condition Mapping.
It is crucial to underscore that \textsf{CTS} is \emph{agnostic} to any specific VAE variant, enabling its seamless integration with a range of leading VAE models.
We will substantiate this property in the generalization test presented in \revise{Sections \ref{sec:expt:fidelity}--\ref{sec:expt:control}}.

\paragraph{Condition Clustering}
This module takes the set $\bm{C}$ of external condition vectors as input and partitions $\bm{X}$ into $k$ clusters, guided by the similarity of their external condition vectors. 
Different types of $k$-means clustering variants are employed depending on the nature of condition values. 
For instance, when dealing with purely categorical condition values--like the binary condition between "periodic" and "not periodic"--we leverage the $k$-modes clustering \cite{huang1998extensions, chaturvedi2001k}.
Conversely, for purely numerical condition values, the $k$-means clustering \cite{macqueen1967some, lloyd1982least, sculley2010web, newling2016nested} is adopted. 
For heterogeneous conditions characterized by a combination of numerical and categorical values, the $k$-prototypes clustering \cite{huang1998extensions} is applied.

After clustering, a representative center is determined for each cluster, such as Mode for $k$-modes clustering and Centroid for $k$-means clustering.
This module yields $k$ clusters, which then facilitate efficient Data Selection during the generation phase, guided by the user-input condition vector $\bm{c_0}$.

\begin{figure*}[t]
\includegraphics[width=0.99\textwidth]{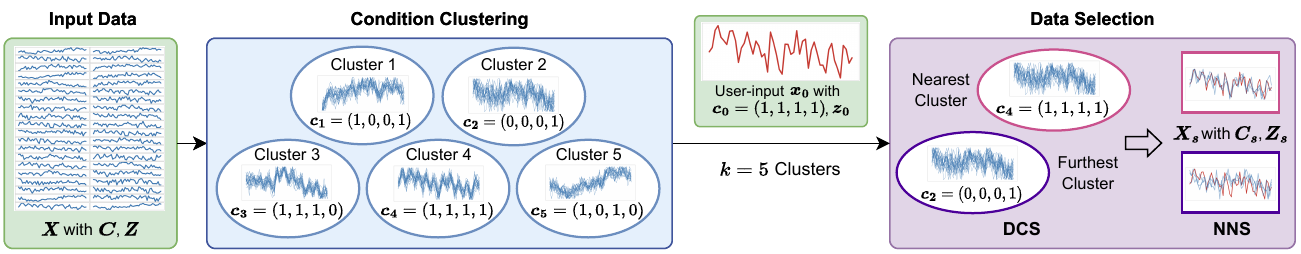}
\vspace{-0.75em}
\caption{An example of Condition Clustering and Data Selection.}
\label{fig:data_selection}
\vspace{-0.25em}
\end{figure*}

\begin{example}
\label{exp:condition_clustering}
  Let us visually illustrate the Condition Clustering module through Figure \ref{fig:data_selection}. 
  Consider a time series dataset characterized by $m=4$ categorical external conditions: \emph{periodicity}, \emph{rapid change}, \emph{increasing/decreasing trend}, and \emph{operating state}. 
  Each condition is indicated by a 1 (present) or 0 (absent) in a time series. 
  For instance, a time series with periodic changes would be annotated with a 1 for "periodicity" and 0 otherwise.
  We employ $k$-modes clustering and set $k=5$, with the result shown in Figure \ref{fig:data_selection}. 
  Each cluster is characterized by a center vector $\bm{c_j}$ ($1 \leq j \leq 5$).
  It reveals that time series sharing similar conditions are aptly grouped into clusters, highlighting the efficacy of Condition Clustering.
  \hfill $\triangle$ \par
\end{example}

\subsection{Generation Phase}
\label{sec:framework:generation}

\noindent\textbf{Data Selection.}
Given a user-input time series $\bm{x_0}$ and its associated external condition vector $\bm{c_0}$, it may be suboptimal to utilize \emph{all} latent vectors $\bm{Z}$ from $\bm{X}$ to learn the mapping function $\bm{f}$. 
This approach risks incorporating data that is dissimilar and has irrelevant conditions, which could introduce noise and ultimately compromise the accuracy and effectiveness of $\bm{f}$.

To empower users to customize different condition vectors $\bm{c_0^\prime}$, we select a subset $\bm{X_s} \subseteq \bm{X}$ with corresponding latent vectors $\bm{Z_s} \subseteq \bm{Z}$, which is chosen to be similar to $\bm{x_0}$ in the latent space yet encompassing a diverse spectrum of external conditions.
This dual objective is achieved through two transparent components: Diverse Condition Selection (\textsf{DCS}) and Nearest Neighbor Search (\textsf{NNS}).

\subparagraph{Diverse Condition Selection (\textsf{DCS})}
Given a parameter $k_1$ and the user-input condition vector $\bm{c_0}$, a trivial approach could be a random selection (\textsf{Rand}), picking $k_1$ clusters uniformly at random to maximize the diversity of external conditions. 
However, \textsf{Rand} may overlook the specifics of $\bm{c_0}$, potentially resulting in selections irrelevant to the user's intended conditions, especially when users typically modify only one or two external conditions.

To strike a balance between relevance and diversity, we select the $\tfrac{k_1}{2}$ closest clusters to $\bm{c_0}$ for direct relevance, and the $\tfrac{k_1}{2}$ furthest clusters to introduce a variety of conditions. 
\textsf{DCS} not only ensures the generated time series closely matches user preferences but also provides a broad spectrum of conditions for further exploration and discovery.
The effectiveness of \textsf{DCS} over \textsf{Rand} will be demonstrated in \revise{Section \ref{sec:expt:attribute}}, highlighting its superior alignment with user preferences and its contribution to exploring diverse scenarios.

\subparagraph{Nearest Neighbor Search (\textsf{NNS})}
This component is tailored to minimize the distance between the latent vector $\bm{z_0}$ of $\bm{x_0}$ and the latent vectors $\bm{z}$ of the time series within the selected $k_1$ clusters from \textsf{DCS}. 
Given another parameter $k_2$, we pinpoint the $k_2$ most similar data for each cluster, culminating in a selection of $(k_1 \cdot k_2)$ time series that constitute the output set $\bm{X_s}$ for Data Selection.

The integrated utilization of \textsf{DCS} and \textsf{NNS} results in a unified way to maximize condition diversity while minimizing the distance in the latent space. This results in a resilient subset $\bm{X_s}$ that enhances the quality of both mapping and generated outcomes.
We will validate this claim through the ablation study in \revise{Section \ref{sec:expt:ablation}}. 

\begin{example}
\label{exp:data_selection}
  We continue to use Figure \ref{fig:data_selection} to elucidate the Data Selection module. We examine the case of a user-input time series $\bm{x_0}$, associated with an external condition vector $\bm{c_0} = (1,1,1,1)$.
  Initially, the \textsf{DCS} component selects two clusters ($k_1 = 2$) based on the diversity of $\bm{c_0}$: Cluster 4 as the nearest and Cluster 2 as the furthest.
  Subsequently, the \textsf{NNS} component identifies three ($k_2 = 3$) most similar time series within Clusters 2 and 4, in terms of closeness to $\bm{x_0}$ in the latent space.
  The synergistic operation of \textsf{DCS} and \textsf{NNS} in the Data Selection module adeptly curates a subset $\bm{X_s} \subseteq \bm{X}$, encompassing a total of six ($\num{\bm{X_s}} = k_1 \cdot k_2 = 6$) time series. 
  This subset skillfully combines visual resemblance to $\bm{x_0}$ with a wide array of external conditions, showcasing the module's proficiency in harmonizing similarity with the user-input time series and diversity in external conditions.
  \hfill $\triangle$ \par
\end{example}

\begin{figure}[h]
\centering
\vspace{-0.5em}
\includegraphics[width=0.99\columnwidth]{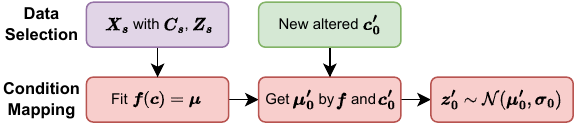}
\vspace{-1.0em}
\caption{The pipeline of Condition Mapping.}
\label{fig:condition_mapping}
\vspace{-0.75em}
\end{figure}

\paragraph{Condition Mapping}
Given a subset $\bm{X_s}$ of time series and their associated external condition vectors $\bm{C_s}$ and latent vectors $\bm{Z_s}$ from Data Selection, we employ a regression model to learn a customized mapping function $\bm{f}$ between $\bm{c} \in \bm{C_s}$ and the mean latent vector $\bm{\mu}$ of $\bm{z} \in \bm{Z_s}$, i.e., $\bm{\mu} = \bm{f}(\bm{c})$. 
This mapping is adopted because $\bm{\mu}$ represents the most significant characteristics of $\bm{x_0}$, enhancing the control over the generated $\bm{x_0^\prime}$ while preserving the essence of $\bm{x_0}$.
The pipeline of Condition Mapping is depicted in Figure \ref{fig:condition_mapping}.

We first train a regression model to learn $\bm{f}$ by a set of pairs $(\bm{c}, \bm{\mu})$ from $\bm{C_s}$ and $\bm{Z_s}$, through minimizing the $l_2$ loss between the true latent vector $\bm{\mu}$ and the predicted latent vector ${\bm{\mu^\prime}}$, i.e., 
\begin{displaymath}
  \bm{f^*} = \textstyle {\argmin}_{\bm{f}} \norm{\bm{\mu} - \bm{\mu^\prime}}_2^2.
\end{displaymath}
When the new altered condition vector $\bm{c_0^\prime}$ is provided, we leverage $\bm{f}$ to determine the mean latent vector $\bm{\mu_0^\prime}$. 
Finally, we obtain the actual latent vector $\bm{z_0^\prime}$ by Gaussian distribution sampling, i.e., $\bm{z_0^\prime} \sim \mathcal{N}(\bm{\mu_0^\prime}, \bm{\sigma_0})$, where $\bm{\sigma_0}$ is the log variance vector corresponding to $\bm{z_0}$, which embodies the inherent uncertainty and variability. 
After Condition Mapping, $\bm{z_0^\prime}$ is passed through the VAE decoder to generate the output time series $\bm{x_0^\prime}$, i.e., $\bm{x_0^\prime} = \bm{p_{\theta}}(\bm{x_0} | \bm{z_0^\prime})$.

Note that the mapping function $\bm{f}$ is learned on the fly during the generation phase.
Compared to conventional strategies that learn a universal $\bm{f}$ from the entire $\bm{X}$ in the training phase, it does require extra time to learn a tailored $\bm{f}$ based on $\bm{X_s}$ and $\bm{C_s}$ for every user-input $\bm{x_0}$ and $\bm{c_0}$. 
Yet, considering that $\num{\bm{X_s}} = k_1 \cdot k_2 \ll \num{\bm{X}} = n$, this learning process remains efficient and lightweight. 
Crucially, it facilitates CTSG across varied external conditions \emph{without} requiring the VAE to undergo retraining.
The practicality and speed of this method will be substantiated in Section \ref{sec:expt:efficiency}.

In real-world applications, we suggest employing \textsf{Decision Tree} \cite{breiman1984cart, quinlan1986induction, quinlan1993c4} as the regression model, driven by two considerations:
(1) As a white-box model \cite{molnar2020interpretable}, \textsf{Decision Tree} provides clear, rule-based decisions, making it highly interpretable and useful for elucidating the control of external conditions.
(2) It excels at capturing intricate, non-linear relationships and demonstrates robustness across different conditions.
A more detailed exploration of different regression models will be presented in \revise{Sections \ref{sec:expt:attribute}--\ref{sec:expt:efficiency}}.
\vspace{-0.25em}

\subsection{Theoretical Analysis}
\label{sec:framework:theory}
After introducing the \textsf{CTS} details, we now pivot to the theoretical analysis and explain the underlying rationale of its module design.
At a high level, users can alter the external conditions via two primary mechanisms: \emph{interpolation} and \emph{extrapolation}. 
Interpolation is the process of estimating a value that lies \emph{within} the known range of training data. 
On the other hand, extrapolation aims to estimate a value \emph{outside} the known range of training data. 
Examples \ref{exp:numerical} and \ref{exp:categorical} elucidate the nuances of interpolation and extrapolation in modifying numerical and categorical conditions, respectively.
\vspace{-0.2em}
\begin{example}[Interpolation and Extrapolation for Numerical Conditions]
\label{exp:numerical}
  Consider a time series air quality dataset with outdoor temperature as a numerical condition. 
  It covers temperatures from 0~$\degree$C to 40~$\degree$C but misses data between 5~$\degree$C and 10~$\degree$C for a specific city.
  For interpolation, users might want to generate a time series at 6~$\degree$C, a missing value within the valid range. 
  For extrapolation, users may target 48~$\degree$C, beyond the dataset's maximum.  
  \hfill $\triangle$ \par
\end{example}
\vspace{-0.5em}
\begin{example}[Interpolation and Extrapolation for Categorical Conditions]
\label{exp:categorical}
  Continuing with the air quality dataset, where weather is a categorical condition.
  It encompasses five weathers: sunny, cloudy, rainy, snowy, and the rare foggy.
  For interpolation, users may want to shift from sunny to the scarcely sampled foggy weather. 
  For extrapolation, the interest could lie in exploring a transformation to Haboob--an intense dust storm not presented in the dataset.
  \hfill $\triangle$ \par
\end{example}
\vspace{-0.25em}

Examples \ref{exp:numerical} and \ref{exp:categorical} illustrate the ubiquity of interpolation and extrapolation when users modify an external condition. 
Next, we systematically demonstrate how \textsf{CTS} supports interpolation and extrapolation for both single and multiple external conditions.
\vspace{-0.2em}
\begin{lemma}[Interpolation for Single Condition]
\label{lemma:interpolation-of-one-condition}
  For a specific external condition within a valid range, \textsf{CTS} can produce coherent outputs for any altered value $c_0^\prime$ within this range.
\end{lemma}
\vspace{-0.25em}
\begin{proof}
  For this proof, as our aim is to modify a single condition value, we employ \emph{scalar representations} for the altered condition and the mapping function. We focus on two types of altered conditions: numerical and categorical.
  
  \subparagraph{Numerical Condition}
  Suppose we have a valid range of $[c_1, c_2]$ for an altered numerical condition, where $c_1 < c_2$. 
  Given a user-input external condition value $c_0$, \textsf{DCS} chooses $k_1$ clusters with diverse conditions, i.e., a half with centers nearest to $c_0$ and a half furthest to $c_0$. Then, we sort them in ascending order.
  For any altered value $c_0^\prime \in [c_1, c_2]$, we use binary search to identify the largest $c_1^\prime$ and the smallest $c_2^\prime$ with the minimum difference to $c_0^\prime$, i.e., $c_1 \leq c_1^\prime \leq c_0^\prime \leq c_2^\prime \leq c_2$. 
  This provides a tight, valid range of $[c_1^\prime, c_2^\prime]$ for $c_0^\prime$.
  
  Employing the mapping function $f$, we get the mean latent values $\mu_1^\prime = f(c_1^\prime)$ and $\mu_2^\prime = f(c_2^\prime)$ for $c_1^\prime$ and $c_2^\prime$, respectively. 
  As $c_0^\prime$ is absent in the training data, a direct computation of its mean latent value $\mu_0^\prime$ using $f$, i.e., $\mu_0^\prime = f(c_0^\prime)$, can introduce significant bias. 
  To reduce bias and generate a coherent output, we derive $\mu_0^\prime$ using linear interpolation \cite{needham1959science, meijering2002chronology} within two valid pairs $(c_1^\prime, \mu_1^\prime)$ and $(c_2^\prime, \mu_2^\prime)$.
  Let $\alpha = (c_0^\prime - c_1^\prime)/(c_2^\prime - c_1^\prime)$, then $\mu_0^\prime$ is computed below: 
  \begin{displaymath}
    \mu_0^\prime = \mu_1^\prime + \alpha \cdot (\mu_2^\prime - \mu_1^\prime).
  \end{displaymath}
  
  \subparagraph{Categorical Condition}
  Suppose we have a valid range of $\{c_1, \cdots, c_v\}$ for an altered categorical condition, with ample training data for the first $(v-1)$ categories but limited data for $c_v$.
  Given a user-input $c_0 \in \{c_1, \cdots, c_{v-1}\}$, learning $f$ for an altered condition $c_0^\prime = c_v$ using all training data may lead to overgeneralization due to data scarcity.
  Yet, \textsf{DCS} effectively addresses this issue by deriving a more pertinent latent value $\mu_0^\prime$ for $c_0^\prime$. \textsf{DCS} acts as a \emph{dual-role} filter:
  \begin{enumerate}[nolistsep,leftmargin=25pt]
    \item It keeps the infrequent category (e.g., $c_v$) by selecting half of the clusters whose centers are the furthest from $c_0$.  
    \item It excludes many irrelevant conditions, retaining only half of the clusters whose centers are the closest to $c_0$.
  \end{enumerate}
  Moreover, \textsf{NNS} ensures $\mu_0^\prime$ maintains the fidelity to input time series.
  Additionally, the VAE's continuous latent space facilitates nuanced and consistent changes in generated data \cite{lyu2022introducing}.
  
  Thus, \textsf{CTS} can produce coherent outputs for any $c_0^\prime$ within a valid range, whether the altered condition is numerical or categorical.
\end{proof}

\vspace{-0.25em}
\begin{corollary}[Interpolation for Multiple Conditions]
\label{corollary:interpolation-of-multi-conditions}
  For multiple external conditions within valid ranges, \textsf{CTS} can produce coherent outputs for any altered vector $\bm{c_0^\prime}$ within these ranges.
\end{corollary}
\vspace{-0.25em}
\begin{sproof}
  For multiple numerical conditions, multivariate interpolation \cite{wiki:mutli-interpolation} is employed to derive $\bm{\mu_0^\prime}$ for any $\bm{c_0^\prime}$ within valid ranges. 
  Given that practical regressors like \textsf{Decision Tree} can discern complex relationships between external conditions and latent features, the proof of Lemma \ref{lemma:interpolation-of-one-condition} can be readily adapted for this corollary.
  To be concise, we omit the details here. 
\end{sproof}

\vspace{-0.25em}
\begin{lemma}[Extrapolation for Single Numerical Condition]
\label{lemma:extrapolation-of-one-condition}
  For a specific numerical external condition within a valid range, \textsf{CTS} can produce coherent outputs for any altered value $c_0^\prime$ outside this range. 
\end{lemma}
\vspace{-0.25em}
\begin{proof}
  We use scalar notations for the altered condition and the mapping function, similar to the proof in Lemma \ref{lemma:interpolation-of-one-condition}. 
  Suppose the valid range of the altered condition is $[c_1, c_2]$, where $c_1 < c_2$. There are two cases for the altered $c_0^\prime$ outside the range of $[c_1, c_2]$:
  \begin{enumerate}[nolistsep,leftmargin=25pt]
    \item $c_2 < c_0^\prime$. 
    For this case, we identify the largest $c_2^\prime$ through binary search from the sorted $k_1$ condition values such that $c_2^\prime \leq c_2 \leq c_0^\prime$. This provides us two nearest values $c_2^\prime$ and $c_2$ to $c_0^\prime$.
    Using $f$, we get $\mu_2^\prime = f(c_2^\prime)$ and $\mu_2 = f(c_2)$. Let $\beta = (c_0^\prime - c_2^\prime)/(c_2 - c_2^\prime)$.
    To reduce bias and produce a coherent output, we derive the mean latent value $\mu_0^\prime$ through linear extrapolation \cite{scott1993causal} within two valid pairs $(c_2^\prime, \mu_2^\prime)$ and $(c_2, \mu_2)$:
    \begin{displaymath}
      \mu_0^\prime = \mu_2^\prime + \beta \cdot (\mu_2 - \mu_2^\prime).
    \end{displaymath}

    \item $c_0^\prime < c_1$.
    Similar to the first case, we can obtain two nearest values $c_1$ and $c_1^\prime$ such that $c_0^\prime \leq c_1 \leq c_1^\prime$ and their mean latent values $\mu_1 = f(c_1)$ and $\mu_1^\prime = f(c_1^\prime)$, respectively.
    Let $\gamma = (c_0^\prime - c_1^\prime)/(c_1 - c_1^\prime)$. Through linear extrapolation  \cite{scott1993causal}, we determine the mean latent value $\mu_0^\prime$ as follows: 
    \begin{displaymath}
      \mu_0^\prime = \mu_1^\prime + \gamma \cdot (\mu_1 - \mu_1^\prime).
    \end{displaymath}
  \end{enumerate}
  
  Thus, \textsf{CTS} can produce a coherent output for any $c_0^\prime$ outside the valid range when the altered condition is numerical.
\end{proof}

\begin{corollary}[Extrapolation for Multiple Numerical Conditions]
\label{corollary:extrapolation-of-multi-conditions}
  For multiple numerical external conditions within valid ranges, \textsf{CTS} can produce coherent outputs for any altered vector $\bm{c_0^\prime}$ outside these ranges.
\end{corollary}
\vspace{-0.5em}
\begin{sproof}
  For multiple numerical conditions, we can use multivariate or geometric extrapolation \cite{scott1993causal, probnet} to derive $\bm{\mu_0^\prime}$ for any $\bm{c_0^\prime}$ outside the valid ranges. Thus, the proof of Lemma \ref{lemma:extrapolation-of-one-condition} can be extended to this scenario.
  We skip the detailed proof for brevity.
\end{sproof}

\vspace{0.2em}
\paragraph{Remarks}
For extrapolation with categorical conditions, if the conditions are \emph{ordinal} with a natural order, $\bm{\mu_0^\prime}$ can be inferred like numerical cases.
On the other hand, for \emph{nominal} data without a specific order, \textsf{CTS} may struggle to produce a reliable $\bm{\mu_0^\prime}$ due to lacking specific external categorical data knowledge, potentially leading to incoherent outputs.
Moreover, while \textsf{CTS} is capable of generating coherent results for numerical conditions in both interpolation and extrapolation, interpolation outcomes are generally more reliable.
This is because interpolation operates with the known data range, whereas extrapolation extends beyond it, with reliability decreasing as the altered value moves further from known data.

\subsection{Time Complexity Analysis}
\label{sec:framework:complexity}

Suppose that there are $n$ training time series, each with $d_r$ raw features, $m$ external conditions, and $d_l$ latent features.
The training phase of \textsf{CTS} has two modules with distinct time complexities:
(1) Data Generation: 
for a VAE model with $l$ hidden layers, each having $h$ neurons and $o$ output neurons, the time complexity of backpropagation is $O(n d_r h^l o e)$, with $e$ being the number of training epochs.
(2) Condition Clustering: 
this module uses a $k$-means style method with a time complexity of $O(nmkt)$, where $t$ is the number of clustering iterations.
Thus, the total time complexity of the training phase is $O(n d_r h^l o e + nmkt)$. It is evident that VAE training is the most time-consuming part of this phase.

The generation phase of \textsf{CTS} also involves two modules, each with distinct time complexities:
(1) Data Selection: this module takes $O(km + \tfrac{n d_l k_1}{k})$ time as
\textsf{DCS} uses $O(km)$ time to select $k_1$ out of $k$ clusters, and
\textsf{NNS} spends $O(\tfrac{n d_l k_1}{k})$ time to identify $(k_1 k_2)$ similar neighbors across $k_1$ clusters, where each cluster has on average $\tfrac{n}{k}$ data.
(2) Condition Mapping: 
learning $\bm{f}$ with \textsf{Decision Tree} incurs $O(k_1 k_2 d_l \log(k_1 k_2))$ time, with inference time being $O(\log(k_1 k_2))$ \cite{breiman1984cart, quinlan1986induction, quinlan1993c4}.
Overall, the generation phase takes $O(km + \tfrac{n d_l k_1}{k} + k_1 k_2 d_l \log(k_1 k_2))$ time. 
As the learning of $\bm{f}$ is independent of $n$, this phase is significantly more efficient than the training phase.

\subsection{Discussions}
\label{sec:framework:discussions}

\noindent\textbf{Versatility.}
\textsf{CTS}, while primarily demonstrated in the time series domain, is versatile to other data modalities due to two main factors:
(1) Its VAE-agnostic nature enables the selection of appropriate VAE models for different data types.
(2) The independence of the mapping function, separated from VAE training, affords exceptional flexibility to its key modules--Condition Clustering, Data Selection, and Condition Mapping across various modalities.
For example, in the image domain, \textsf{CTS} can be extended for controllable image generation by employing leading supervised disentangled VAEs such as \textsf{DC-VAE} \cite{parmar2021dual}, \textsf{Soft-IntroVAE} \cite{daniel2021soft} and \textsf{VQ-VAE} \cite{vqvae} to produce the generated images. 
The versatility of \textsf{CTS} across different data modalities will be showcased in \revise{Section \ref{sec:expt:versatility}}.

\paragraph{Explainability}
In the \textsf{CTS} framework, the Data Selection module utilizes two transparent components (i.e., \textsf{DCS} and \textsf{NNS}) to pinpoint similar time series across diverse external conditions. 
Moreover, the Condition Mapping module employs the white-box regression model (e.g., \textsf{Decision Tree}) to effectively map latent features to these conditions. 
Together, these modules ensure that the generated outputs not only align with user preferences but also offer enhanced explainability. 
An illustrative case study demonstrating this capability will be presented in \revise{Section \ref{sec:expt:explanability}}.

%% file: 07_evaluation.tex
\section{CTSG Evaluation Scheme}
\label{sec:eval}

\begin{figure}[h]
\centering
\vspace{-0.5em}
\includegraphics[width=0.99\columnwidth]{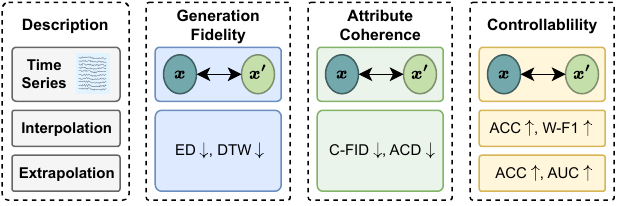}
\vspace{-1.0em}
\caption{CTSG evaluation scheme.}
\label{fig:eval}
\vspace{-1.0em}
\end{figure}

\noindent\textbf{Motivation.}
Current evaluations of controllable generation typically concentrate on the degree of disentanglement \cite{deng2020disentangled, chen2018isolating, kim2018disentangling}. 
While these measures reveal how latent features vary under different external conditions, they often overlook the detailed interplay between these features and conditions.
Moreover, many disentangled VAEs sacrifice generation quality for greater disentanglement \cite{shao2022rethinking}, neglecting a thorough assessment of generated data quality and lacking in both data fidelity and controllability.
To bridge these gaps, we introduce a comprehensive CTSG evaluation scheme that focuses on three aspects: generation fidelity, attribute coherence, and controllability, as presented in Figure \ref{fig:eval}.

\paragraph{Generation Fidelity} 
A proficient CTSG method ought to generate high-quality data that \emph{align closely with the input data}.  
For evaluating this objective, we employ two measures widely recognized in TSG tasks and assessment \cite{tsgbench}, i.e., Euclidean Distance (ED) and Dynamic Time Warping (DTW) \cite{dtw}.
They are applied for both interpolation and extrapolation scenarios, with lower values signifying enhanced fidelity to the original time series.

\paragraph{Attribute Coherence}
A viable CTSG method should also \emph{preserve the inherent attributes of input data} without introducing unintended distortions.
To measure it, we utilize Contextual-Frechet Inception Distance (C-FID) \cite{psa-gan, tt-aae} and AutoCorrelation Difference (ACD) \cite{lai2018modeling} in both interpolation and extrapolation contexts.
They are crucial in determining the extent to which the generated $\bm{x^\prime}$ resembles the input $\bm{x}$, with lower scores reflecting a stronger preservation of inherent attributes.
In our experiments, we compute the mean values by comparing $\bm{x^\prime}$ with $\bm{x}$ using a random sample set, providing a robust and convincing assessment of the CTSG performance. 

\paragraph{Controllability} 
To accurately gauge the quality of controllability, we aim to assess how well \emph{condition alterations meet the intended outcomes}. 
Given that most time series datasets do not have ground truths, the accurate recognition of input patterns becomes essential for distinguishing between existing and generated conditions.
Thus, we employ distinct tasks and corresponding measures tailored for interpolation and extrapolation scenarios.
\begin{enumerate}[nolistsep,leftmargin=25pt]
  \item \keypoint{Interpolation} 
  We employ a multi-class classifier for interpolation based on \textsf{ROCKET} \cite{dempster2020rocket}.  
  It is trained on $\bm{x}$, validated on a held-out set, and then tested on the generated $\bm{x}^\prime$. 
  Performance is gauged using Accuracy (ACC) and Weighted-F1 (W-F1). 
  Ideally, the generated time series should resemble the validation results, thereby affirming the classifier's ability to differentiate between classes.

  \item \keypoint{Extrapolation} 
  We utilize an anomaly detector based on \textsf{ECOD} \cite{li2022ecod}, leveraging ACC and Area Under the Curve (AUC) as evaluation metrics. 
  This detector, trained on $\bm{x}$ and validated on a held-out set, is designed to detect shifts from normal to extreme conditions. 
  High-quality generated time series $\bm{x}^\prime$ should be flagged as outliers, potentially matching or even surpassing the validation set's performance.
\end{enumerate}

\paragraph{Remarks}
This evaluation scheme is flexibly designed to accommodate various data modalities, especially those lacking ground truth.
Next, we will employ this evaluation scheme to assess the performance of \textsf{CTS} for both time series and image domains.

%% file: 08_experiments.tex
\section{Experiments}
\label{sec:expt}

In this section, we showcase the efficacy of \textsf{CTS}, focusing on addressing the following research questions:
\begin{itemize}[nolistsep,leftmargin=20pt]
  \item \textbf{Generation Quality and Attribute Coherence:} 
  Does \textsf{CTS} achieve high fidelity in generation and preserve essential patterns and attributes? (Sections \ref{sec:expt:fidelity} and \ref{sec:expt:attribute})
  
  \item \textbf{Controllability:} 
  Can \textsf{CTS} adeptly support controllability through interpolation and extrapolation? (Section \ref{sec:expt:control})

  \item \textbf{Efficiency:} 
  What are the efficiency and scalability of \textsf{CTS}, especially in generating user-altered conditions? (Section \ref{sec:expt:efficiency})

  \item \textbf{Ablation Study:} 
  What is the impact of \textsf{DCS} and \textsf{NNS} within the Data Selection module in \textsf{CTS}, and how do these components contribute to its effectiveness? (Section \ref{sec:expt:ablation})

  \item \textbf{Sensitivity:}
  How do the key parameters in \textsf{CTS} affect its performance? (Section \ref{sec:expt:para})
  
  \item \textbf{Versatility:}
  Can \textsf{CTS} be effectively applied across different modalities, and how does it perform compared to other advanced methods in these settings? (Section \ref{sec:expt:versatility})
  
  \item \textbf{Explainability:} How transparent are the intermediate outputs and generation results from \textsf{CTS} to users? (Section \ref{sec:expt:explanability})
\end{itemize}
Before delving into the results, we detail the experimental setup.

\subsection{Experimental Setup}
\label{sec:expt:setup}

\noindent\textbf{Datasets.}
We adopt three real-world time series datasets in our experiments: Air Quality Guangzhou (\textbf{Air-GZ}) \cite{zheng2015forecasting}, Air Quality Shenzhen (\textbf{Air-SZ}) \cite{zheng2015forecasting}, and \textbf{Boiler} \cite{sasa}. 
To make a systematic comparison, we exclude data without or lacking the selected conditions to guarantee distinct and observable changes.

\paragraph{Benchmark Methods}
In the absence of direct competitors for the new CTSG problem, we have devised several variants of \textsf{CTS} for comparative analysis, focusing on two essential aspects:
\begin{enumerate}[nolistsep,leftmargin=25pt]
  \item \keypoint{Condition Selection} 
  We examine two strategies for selecting external conditions in the Data Selection module: \textsf{Rand} and \textsf{DCS}, as detailed in Section \ref{sec:framework:generation}.
  
  \item \keypoint{Condition Mapping Function} 
  We explore three different functions in the Condition Mapping module: \textsf{Linear Regression} (\textsf{LR}), \textsf{Random Forest} (\textsf{RF}), and \textsf{Decision Tree} (\textsf{DT}).
\end{enumerate}

This leads to six distinct methods for benchmarking: \textbf{\textsf{Rand-LR}}, \textbf{\textsf{Rand-RF}}, \textbf{\textsf{Rand-DT}}, \textbf{\textsf{DCS-LR}}, \textbf{\textsf{DCS-RF}}, and \textbf{\textsf{DCS-DT}} (the standard \textbf{\textsf{CTS}}).
For these \textsf{CTS} variants, we use \textbf{\textsf{TimeVAE}} \cite{timevae}, an advanced VAE model for TSG. We set $d_l = 16$ for Air-GZ, $d_l = 8$ for Air-SZ, and $d_l = 32$ for Boiler.
Furthermore, to evaluate the VAE-agnostic nature of \textsf{CTS}, we also employ \textbf{\textsf{TimeVQVAE}} \cite{timevqvae} for TSG (referring to this variant as \textbf{\textsf{CTS}$_{\textsf{VQ}}$}) and conduct generalization tests. 
Following the guidelines of \textsf{TimeVQVAE}, we set n\_fft $=8$ and max\_epochs $\in \{2000, 10000\}$ for its two training stages.

\paragraph{Evaluation Measures}
We employ the evaluation scheme outlined in Section \ref{sec:eval} to systematically assess different CTSG methods.

\paragraph{Implementation and Environment}
All experiments are conducted on a machine with Intel\textsuperscript{\textregistered} Core\textsuperscript{\textregistered} i5-13600KF, 32 GB memory, and an NVIDIA RTX 4090, running on Ubuntu 22.04.

\begin{figure}[h]
\centering
\vspace{-0.25em}
\includegraphics[width=0.99\columnwidth]{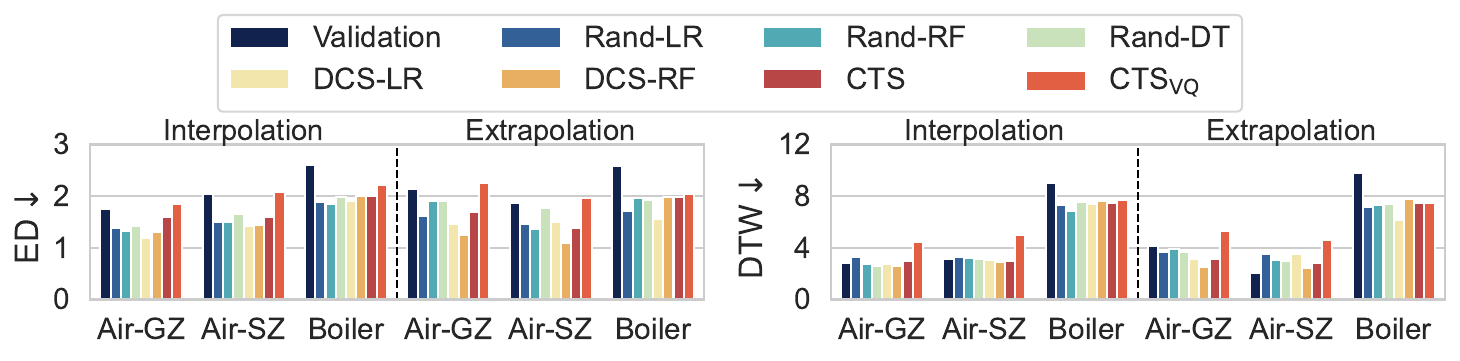}
\vspace{-1.0em}
\caption{Generation fidelity results on time series datasets.}
\label{fig:ts_fidelity}
\vspace{-1.25em}
\end{figure}

\subsection{Generation Fidelity}
\label{sec:expt:fidelity}

\noindent\textbf{Interpolation and Extrapolation.}
The results depicted in Figure \ref{fig:ts_fidelity} reveal that \textsf{CTS} delivers excellent results in both ED and DTW across interpolation and extrapolation. 
This consistent performance highlights \textsf{CTS}'s capability to closely mimic the original time series in terms of both value accuracy and trend alignment.
This enhanced performance can be attributed to (1) the robust generation capability of the Data Generation module, ensuring little diminution in the distinctiveness of generated series, and (2) the negligible impact of other \textsf{CTS} modules on the generation fidelity.

\paragraph{Generalization Test}
From Figure \ref{fig:ts_fidelity}, the ED and DTW values of \textsf{CTS}$_{\textsf{VQ}}$ are generally on par with, or slightly less optimal than, those of \textsf{CTS}.
However, \textsf{CTS}$_{\textsf{VQ}}$ still outperforms most of validation sets. 
This outcome supports the assertion made in Section \ref{sec:framework:train} about \textsf{CTS}'s adaptability to various VAE architectures.

\subsection{Attribute Coherence}
\label{sec:expt:attribute}

\begin{figure}[t]
\centering
\includegraphics[width=0.99\columnwidth]{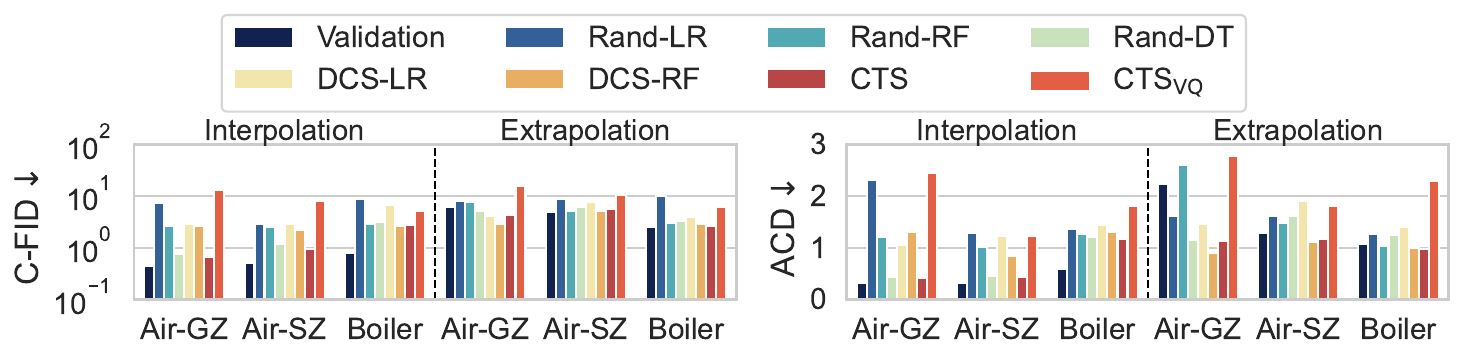}
\vspace{-1.0em}
\caption{Attribute coherence results on time series datasets.}
\label{fig:ts_attribute}
\vspace{-0.25em}
\end{figure}

\noindent\textbf{Interpolation and Extrapolation.}
Figure \ref{fig:ts_attribute} displays the attribute coherence results.
\textsf{CTS} records C-FID and ACD scores that are not only comparable to validation sets but also outperform most competitors.
Thus, the time series generated by \textsf{CTS} effectively mirror the original ones in inherent attributes, confirming their suitability for further analytical tasks.
Compared to \textsf{Rand}, \textsf{DCS} exhibits greater efficacy in improving attribute coherence. 
As discussed in Section \ref{sec:framework:generation}, this superiority stems from \textsf{DCS}'s comprehensive coverage of diverse conditions, facilitating the generation of time series that more precisely reflect user-specified parameters.
For Condition Mapping, the methods based on \textsf{DT} and \textsf{RF} outperform the ones with \textsf{LR}. 
This superiority is due to their more adept handling of complex interrelations between external conditions and latent features, yielding a more refined and effective mapping process.

\paragraph{Generalization Test}
According to Figure \ref{fig:ts_attribute}, \textsf{CTS}$_{\textsf{VQ}}$ exhibits diminished performance compared to \textsf{CTS}.
The performance variation in \textsf{CTS}$_{\textsf{VQ}}$ can be ascribed to the inherent trade-offs in the latent space configuration of \textsf{TimeVQVAE}.
When utilizing a high-dimensional latent space, the mapping function's performance tends to deteriorate, likely due to the increased complexity and potential overfitting issues. Conversely, opting for a low-dimensional latent space results in insufficient representation capacity, leading to the loss of key inherent attributes, as the reduced space cannot adequately encapsulate all the necessary information. 
Nevertheless, \textsf{CTS}$_{\textsf{VQ}}$ still maintains a level of attribute coherence that is commendable, underscoring the versatility and effectiveness of the \textsf{CTS} framework across different VAE models.

\input{tables/control_ts}

\vspace{-0.9em}
\subsection{Controllability}
\label{sec:expt:control}
\noindent\textbf{Interpolation.}
Table \ref{tab:control-time-series} presents the controllability results.
For the interpolation, \textsf{CTS} demonstrates superior performance on most datasets, outperforming other benchmark methods. 
This is primarily attributed to the effective integration of the \textsf{DCS} and \textsf{NNS} components and the condition mapping function using \textsf{DT}.
A notable observation is the performance gap between \textsf{Rand} and \textsf{DCS}. \textsf{Rand} falls short in effectively choosing relevant samples, which is crucial for high-quality generation.
Moreover, the methods employing \textsf{DT} and \textsf{RF} as the condition mapping function outperform those using \textsf{LR}.
This indicates \textsf{LR}'s limitations in capturing the complex dynamics involved in the control process.

\paragraph{Extrapolation}
Turning to extrapolation, especially when detecting abnormal time series under extreme conditions, \textsf{CTS} excels by frequently achieving higher ACC and AUC values than the validation sets. 
This notable achievement, when coupled with its exceptional generation fidelity, indicates that \textsf{CTS} adeptly meets two critical objectives: (1) adherence to user-specified conditions, particularly in rare and extreme scenarios, and (2) the retention of key characteristics from historical data.
This efficacy of \textsf{CTS} is largely due to its \textsf{DCS} component, which adeptly selects conditions closely related to the modified ones. 
The \textsf{NNS} component further ensures that the generated series retain crucial characteristics, aligning precisely with user-defined conditions.
These capabilities allow \textsf{CTS} to handle a range of condition alterations, from ordinary to extreme, ensuring both high fidelity and relevance in the generated series, especially in scarce data scenarios.

\paragraph{Generalization Test}
The results in Table \ref{tab:control-time-series} also show a reduced performance of \textsf{CTS}$_{\textsf{VQ}}$ compared to \textsf{CTS}.
This is linked to \textsf{TimeVQVAE}'s focus on data fidelity over effectively mapping latent features to external conditions, which is crucial for controllable generation.
However, in extrapolation tasks involving rare or extreme conditions, \textsf{CTS}$_{\textsf{VQ}}$ approximately matches \textsf{CTS}'s performance, demonstrating its strength in handling scarce data scenarios.

\begin{figure}[ht]
\centering
\vspace{-1.0em}
\captionsetup[subfigure]{belowskip=-10.0em}
\subfigure[Time Series Generation.]{%
  \label{fig:efficiency:1}%
  \includegraphics[width=0.49\columnwidth]{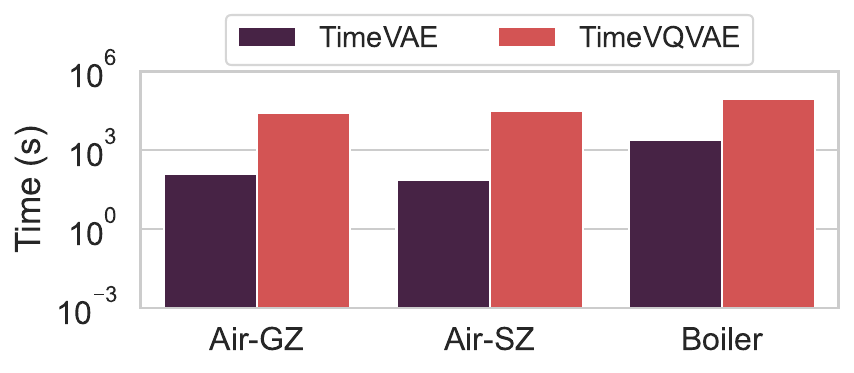}}%
\subfigure[Condition Clustering.]{%
  \label{fig:efficiency:2}%
  \includegraphics[width=0.49\columnwidth]{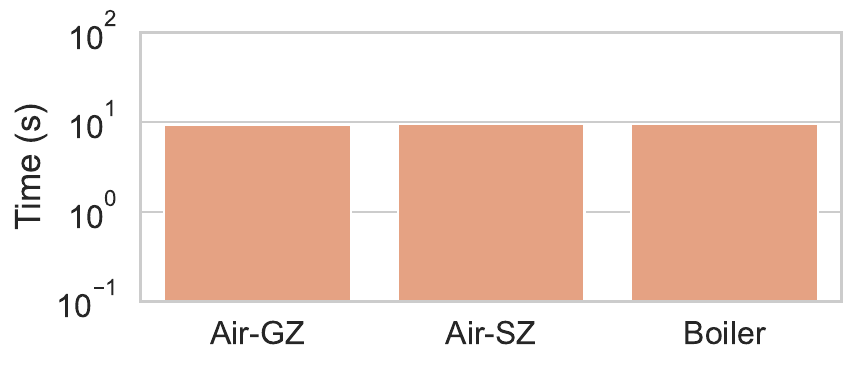}} \\
\vspace{-0.75em}
\subfigure[Data Selection.]{%
  \label{fig:efficiency:3}%
  \includegraphics[width=0.49\columnwidth]{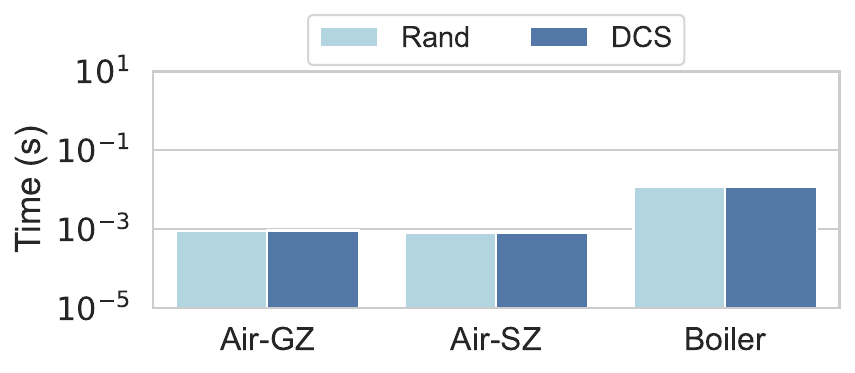}}%
\subfigure[Condition Mapping.]{%
  \label{fig:efficiency:4}%
  \includegraphics[width=0.49\columnwidth]{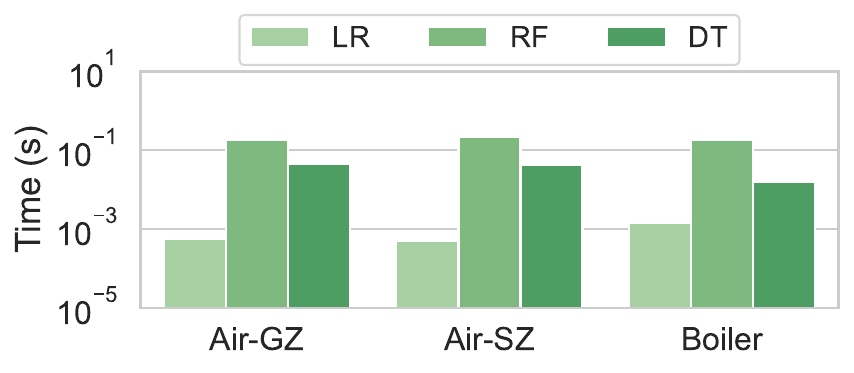}}%
\vspace{-1.5em}
\caption{Running time of different modules in \textsf{CTS}.}
\label{fig:efficiency}
\vspace{-1.25em}
\end{figure}

\subsection{Efficiency}
\label{sec:expt:efficiency}

Figure \ref{fig:efficiency} details the running time of each module in \textsf{CTS}. 
In the Data Generation module, \textsf{TimeVAE} operates at least an order of magnitude faster than \textsf{TimeVQVAE} (Figure \ref{fig:efficiency:1}). 
Yet, due to the complexity of VAE training, Data Generation is significantly slower than other \textsf{CTS} modules, taking at least two orders of magnitude longer (Figures \ref{fig:efficiency:2}--\ref{fig:efficiency:4}).
Figure \ref{fig:efficiency:3} reveals that Data Selection is the most efficient among all four modules. This efficiency arises from its simple yet effective methods (\textsf{Rand} and \textsf{DCS}), which rely primarily on specified conditions.
Further, the time for Condition Mapping is largely influenced by the choice of regression model (Figure \ref{fig:efficiency:4}). The ensemble-based \textsf{RF} model, for instance, demands more processing time than simpler options like \textsf{DT} and \textsf{LR}.

Moreover, we observe that for datasets with a growing number of samples ($n$) (e.g., from Air-SZ to Boiler), there is an increase in the running time for both Data Generation and Data Selection. 
Yet, the time of other modules remains largely unaffected, as their operations are less sensitive to $n$. This behavior underscores the scalability of \textsf{CTS}. 
These empirical observations are consistent with the time complexity analysis detailed in Section \ref{sec:framework:complexity}.

Figures \ref{fig:efficiency:3} and \ref{fig:efficiency:4} also show that the average time taken by \textsf{CTS} to modify external conditions is about 8.1 milliseconds. 
This response time falls well within the Human-Computer Interaction standard of 200 milliseconds, which is typically imperceptible to users \cite{kohrs2016delays, gutwin2015testing}. 
Such rapid responsiveness substantiates the efficiency of \textsf{CTS}, as claimed in Section \ref{sec:framework:generation}. 
Thus, \textsf{CTS} produces high-quality CTSG results and operates efficiently for a seamless user experience.

\subsection{Ablation Study}
\label{sec:expt:ablation}

We conduct an ablation study to evaluate the impact of \textsf{DCS} and \textsf{NNS} within the Data Selection module of \textsf{CTS}. 
To enable a thorough comparison, we introduce three variants of \textsf{CTS}: 
(1) \textsf{CTS$-$NNS}, which excludes \textsf{NNS}; 
(2) \textsf{CTS$-$DCS}, which omits \textsf{DCS}; and 
(3) \textsf{CTS$-$NNS$-$DCS}, which operates without Data Selection.

\begin{figure}[t]
\centering
\includegraphics[width=0.99\columnwidth]{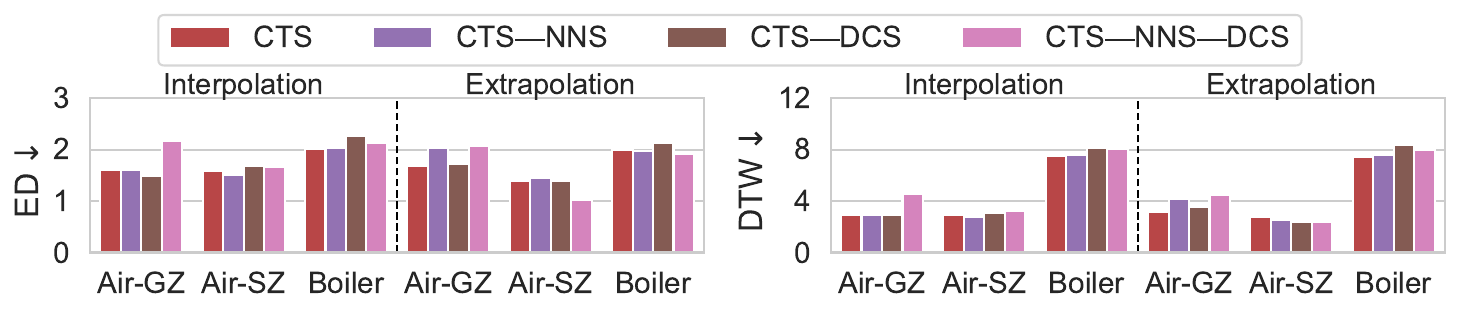}
\vspace{-1.0em}
\caption{Ablation study of generation fidelity.}
\label{fig:ts_fidelity_ablation}
\vspace{-0.5em}
\end{figure}

\paragraph{Generation Fidelity}
Figure \ref{fig:ts_fidelity_ablation} presents the generation fidelity results. 
\textsf{CTS} registers the lowest ED and DTW values across different datasets and scenarios (both interpolation and extrapolation). This performance indicates its effectiveness in generating time series that closely resemble the original data regarding key patterns.
In particular, \textsf{CTS$-$NNS} generally yields inferior results. This disparity highlights the critical role of \textsf{NNS} in enhancing the generation fidelity of \textsf{CTS}, as it effectively filters out irrelevant samples from historical data.

\begin{figure}[t]
\centering
\includegraphics[width=0.99\columnwidth]{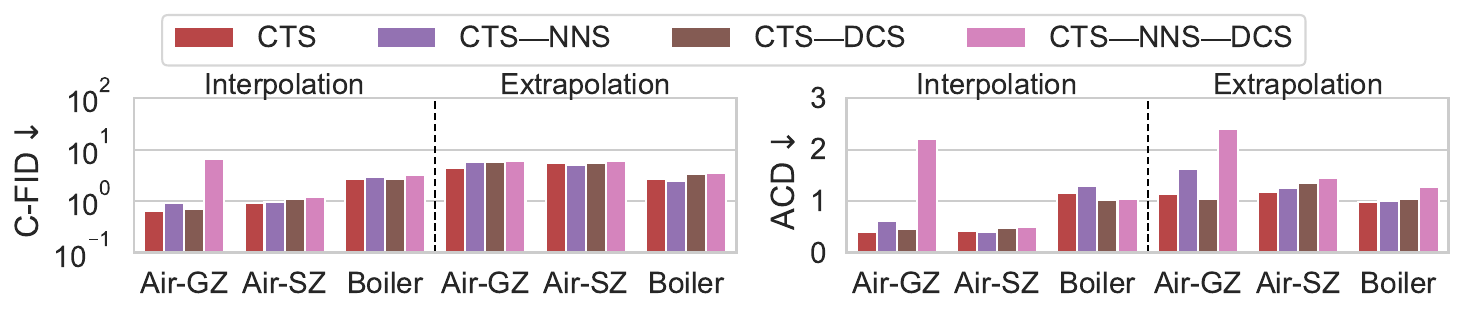}
\vspace{-1.0em}
\caption{Ablation study of attribute coherence.}
\label{fig:ts_attribute_ablation}
\vspace{-0.25em}
\end{figure}

\paragraph{Attribute Coherence}
As evidenced in Figure \ref{fig:ts_attribute_ablation}, \textsf{CTS} achieves the lowest C-FID and ACD values, and \textsf{CTS$-$NNS} and \textsf{CTS$-$DCS} have lower C-FID and ACD values than \textsf{CTS$-$NNS$-$DCS} for all datasets.
The findings highlight that both \textsf{DCS} and \textsf{NNS} components are individually instrumental in preserving the attribute coherence of the generated series to input data.
When they are employed together, they synergize to significantly enhance the quality of both mapping and generated outcomes, as discussed in Section \ref{sec:framework:generation}.

\input{tables/ablation_control}

\paragraph{Controllability}
Table \ref{tab:ablation-control} depicts the controllability results. \textsf{CTS} significantly improves the controllability compared to \textsf{CTS$-$NNS$-$DCS}.
For instance, \textsf{CTS} surpasses \textsf{CTS$-$NNS$-$DCS} in interpolation by an average of 42.4\%, and in extrapolation by 16.9\%. 
Such results highlight the effectiveness of the Data Selection component in \textsf{CTS}, which adeptly selects samples with diverse conditions, thereby facilitating improved CTSG.
Furthermore, the gains in performance with \textsf{CTS} are comparatively modest in the Boiler dataset than in Air-GZ and Air-SZ. This can be attributed to the more complex features in Boiler, making Data Selection more challenging to capture all key patterns.
Despite this, \textsf{CTS} consistently excels in all metrics. This consistency across various datasets reinforces the value of Data Selection in \textsf{CTS}, particularly in enhancing controllability in TSG, even in data with complex patterns.

\subsection{Sensitivity}
\label{sec:expt:para}

\begin{figure}[t]
\centering
\subfigure[Sensitivity of $k$ (1\textsuperscript{st} row: interpolation, 2\textsuperscript{nd} row: extrapolation).]{
  \label{fig:para-k-boiler}
  \includegraphics[width=0.99\columnwidth]{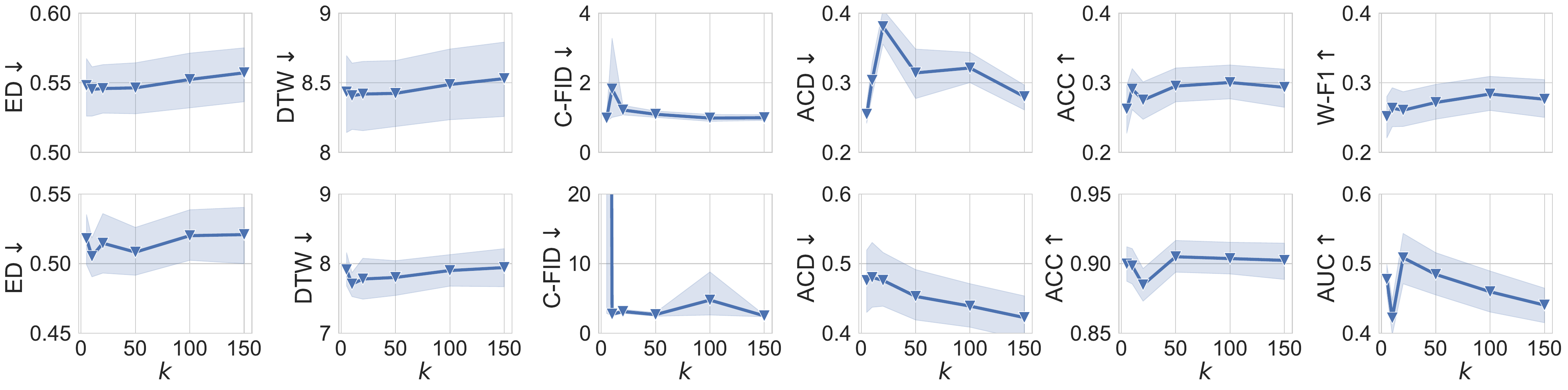}} \\
\vspace{-0.5em}
\subfigure[Sensitivity of $k_1$ and $k_2$ (1\textsuperscript{st} row: interpolation, 2\textsuperscript{nd} row: extrapolation).]{
  \label{fig:para-k1k2-boiler}
  \includegraphics[width=0.99\columnwidth]{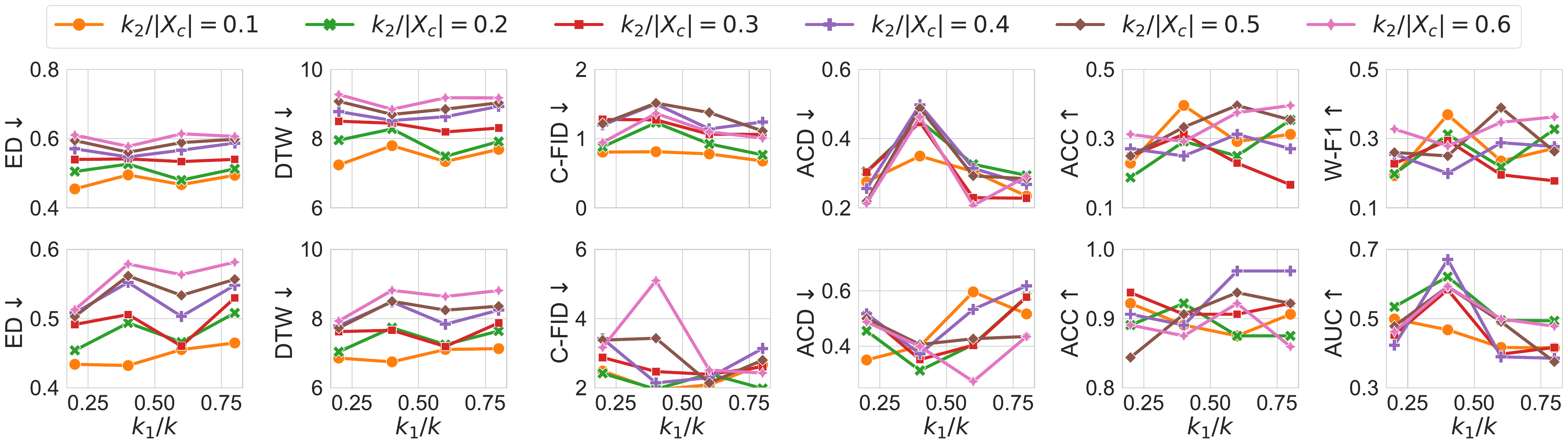}} \\
\vspace{-1.25em}
\caption{Sensitivity of $k$, $k_1$, and $k_2$ in \textsf{CTS} on Boiler.}
\label{fig:para}
\end{figure}

We further study the sensitivity of three key parameters in \textsf{CTS}, i.e., the number $k$ of clusters, the number $k_1$ of selected clusters, and the number $k_2$ of the most similar time series in each selected cluster. 
The results on the Boiler dataset are depicted in Figure \ref{fig:para}. Similar trends can be observed from Air-GZ and Air-SZ.

\paragraph{Sensitivity of $k$}
We experiment with $k$ values in a set of $\{5, 10, 20,$ $50, 100, 150\}$, examining all possible combinations of $k_1$ and $k_2$. The results are shown in Figure \ref{fig:para-k-boiler}.
A notably large $k$ (e.g., $k>100$) enhances controllability but may compromise generation fidelity due to excessively granular clusters.
In contrast, a small $k$ (e.g., $k<20$) leads to too diverse clusters, which can hinder Data Selection and affect attribute coherence. 
Thus, a $k$ value of 50 generally offers the most effective performance in our experiments.

\paragraph{Sensitivity of $k_1$ and $k_2$}
We set $k=50$ and visualize the results by varying the ratio of $k_1$ to $k$ (i.e., $k_1/k$) and the ratio of $k_2$ to $X_c$ (i.e., $k_2/|X_c|$), where $|X_c|$ is the average number of time series in each cluster. 
Specifically, we consider $k_1/k$ in a set of $\{0.2, 0.4, 0.6, 0.8\}$ and $k_2/|X_c|$ within the range of $[0.1, 0.6]$ as a very large value of $k_2/|X_c|$ would inadvertently incorporate not-so-close neighbors. The results are depicted in Figure \ref{fig:para-k1k2-boiler}. 

For a small value of $k_2/|X_c|$ (e.g., $k_2/|X_c| \leq 0.2$), the results are unstable and often lead to worse performance. 
For example, both attribute coherence and controllability metrics show a decline with such settings.
This instability is due to the nearest neighbors' limited diversity, hampering the decoder's generative capabilities. 
Hence, a moderate value for $k_2/|X_c|$ span of $[0.4, 0.6]$ is recommended.
We then focus on $k_2/|X_c| \in [0.4, 0.6]$. 
For $k_1/k$, larger values (e.g., $k_1/k \geq 0.4$) typically yield better results due to the inclusion of more diverse conditions, facilitating the generation of more representative data.
Yet, the performance tends to deteriorate when $k_1/k$ becomes excessively large (e.g., $k_1/k > 0.6$).
This decline emerges from selecting too many dissimilar conditions unrelated to the input time series, thereby compromising the generation quality. Thus, a more balanced $k_1/k \in [0.4, 0.6]$ is preferred.

\subsection{Versatility}
\label{sec:expt:versatility}

\noindent\textbf{Configuration.}
To showcase the versatility of \textsf{CTS}, we extend its application to the image domain.
We utilize two facial datasets without ground truth (\textbf{CelebA} \cite{liu2015deep} and \textbf{FFHQ} \cite{karras2019style}), and one structural dataset with ground truth (\textbf{Teapots} \cite{eastwood2018a}) to ensure comprehensive benchmarking.
We resize all images to a resolution of 32$\times$32 and normalize condition values to fall in a range of $[0,1]$ for consistency in comparison.
We select three state-of-the-art supervised disentangled VAE models as competitors, namely \textbf{\textsf{CCVAE}} \cite{joy2021capturing}, \textbf{\textsf{Guided-VAE}} \cite{ding2020guided} and \textbf{\textsf{IDVAE}} \cite{mita2021identifiable}, known for their ability to accurately reflect changes in user-defined conditions and generate high-quality images.
For \textsf{CCVAE} \cite{joy2021capturing}, we adopt its optimal semi-supervised settings with $d_l = 45$.
For \textsf{Guided-VAE} \cite{ding2020guided}, we adhere to the authors' advice for multi-dimensional conditions.
For \textsf{IDVAE} \cite{mita2021identifiable}, we fine-tune the model based on its code examples to achieve optimal results.
For \textsf{CTS}, we use \textbf{\textsf{DC-VAE}} \cite{parmar2021dual} for image generation, setting $d_l = 128$ and the learning rate to 0.0002.

\begin{figure}[t]
\centering
\includegraphics[width=0.99\columnwidth]{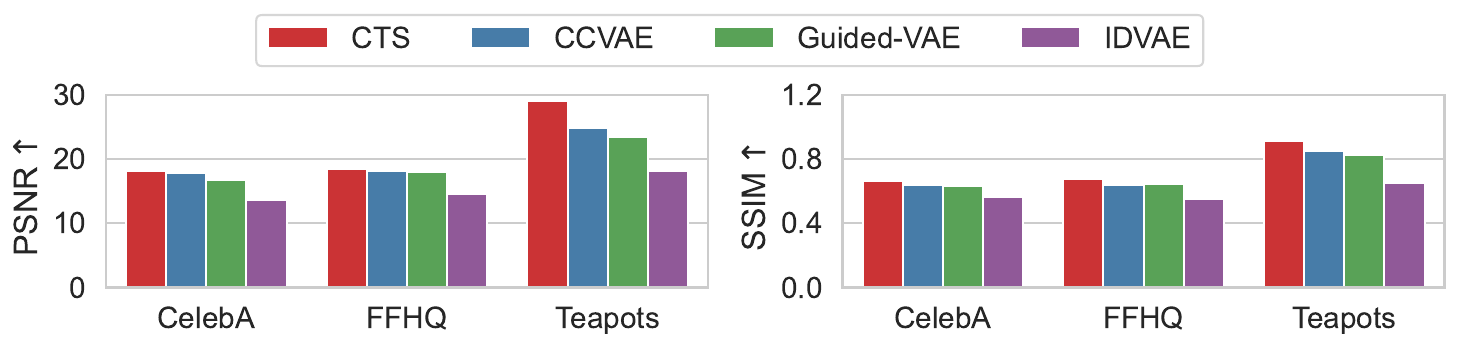}
\vspace{-1.0em}
\caption{Generation fidelity results on image datasets.}
\label{fig:image_fidelity}
\vspace{-0.5em}
\end{figure}

\paragraph{Generation Fidelity} 
We first use PSNR and SSIM \cite{wang2004image} to measure the fidelity of generated images $\bm{x^\prime}$ to their original ones $\bm{x}$. 
The results are shown in Figure \ref{fig:image_fidelity}.
\textsf{CTS} excels over the three competitors in PSNR and SSIM across all datasets, showcasing its superior generation fidelity. 
This efficacy arises from \textsf{DC-VAE}'s ability to discern intricate dependencies between external conditions, 
while disentangled VAEs, in their quest for latent feature independence, may inadvertently compromise generation fidelity.

\begin{figure}[t]
\centering
\includegraphics[width=0.99\columnwidth]{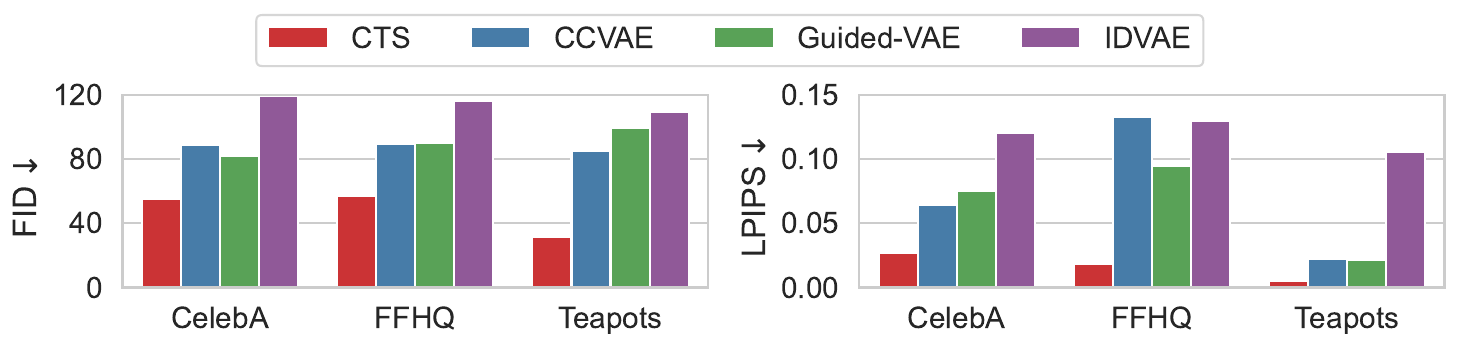}
\vspace{-1.0em}
\caption{Attribute coherence results on image datasets.}
\label{fig:image_attribute}
\vspace{-0.25em}
\end{figure}

\paragraph{Attribute Coherence}
We then adopt FID \cite{heusel2017gans} and LPIPS \cite{zhang2018unreasonable} to evaluate the attribute coherence between $\bm{x^\prime}$ and $\bm{x}$. 
The results are visualized in Figure \ref{fig:image_attribute}.
\textsf{CTS} consistently achieves lower FID and LPIPS than its competitors, underlining its ability to generate images that are perceptually akin to the original under altered conditions.
These results further confirm that \textsf{CTS} can better preserve the inherent characteristics and generate more visually similar images than advanced disentangled VAEs.

\input{tables/control_image}

\paragraph{Controllability}
For CelebA and FFHQ, which lack ground truth, we follow \cite{kowalski2020config} and use $I_+$, $I_-$, and MD to train an attribute predictor to estimate whether a condition is altered desirably. 
For Teapots, the controllability is evaluated using PSNR and SSIM metrics, leveraging ground truth images for precise assessment. 
The outcomes are detailed in Table \ref{tab:control-image}.
\textsf{CTS} either surpasses or matches the performance of the three leading disentangled VAEs, highlighting its strong capability to generate images with user-desired conditions.
For MD, \textsf{CTS} exhibits slightly larger values than the three competitors. 
This is largely due to the disentangled VAEs' ability to learn relationships between independent latent features and external conditions, while \textsf{CTS} is tailored to manage more complex relationships. Thus, \textsf{CTS} can generate images with desired conditions without compromising the quality of other features. 

\input{tables/generalization_test_image}

\paragraph{Generalization Test}
Unlike the time series domain, \textsf{CTS} can integrate a variety of advanced VAE models for image generation. To validate its adaptability with different VAEs, we conduct generalization tests using two notable VAEs: \textbf{\textsf{Soft-IntroVAE}} \cite{daniel2021soft} and \textbf{\textsf{VQ-VAE}} \cite{vqvae}.
The results in Table \ref{tab:generalization-image} show that the choice of VAE minimally impacts \textsf{CTS}'s generation quality and controllability. 
Notably, \textsf{CTS} consistently surpasses three competitors in most scenarios, even with the less optimal \textsf{VQ-VAE}, as shown in Figure \ref{fig:image_fidelity}, Figure \ref{fig:image_attribute}, and Table \ref{tab:control-image}.
These results further affirm that \textsf{CTS} is VAE-agnostic, capable of leveraging various leading VAEs across different data domains for effective data generation.

\begin{figure*}[t]
\centering
\vspace{-0.5em}
\subfigure[Visualization of the Data Selection module.]{
  \label{fig:case_ffhq}
  \includegraphics[width=0.99\textwidth]{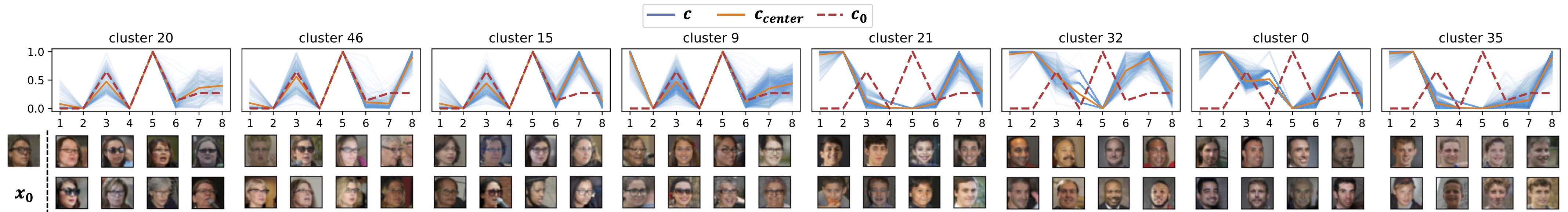}}
\vspace{-0.25em}
\subfigure[Visualization of the Condition Mapping module through \textsf{Decision Tree}.]{
  \label{fig:dt}
  \includegraphics[width=0.99\textwidth]{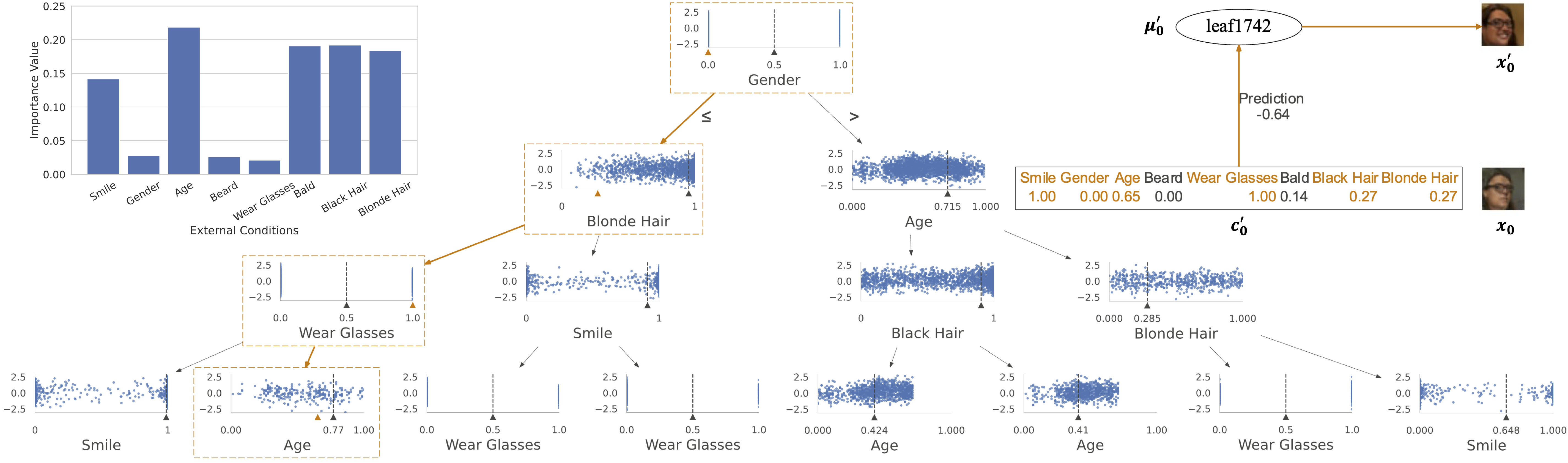}}
\vspace{-0.25em}
\subfigure[Latent traversals.]{
  \label{fig:case_walk}
  \includegraphics[width=0.99\textwidth]{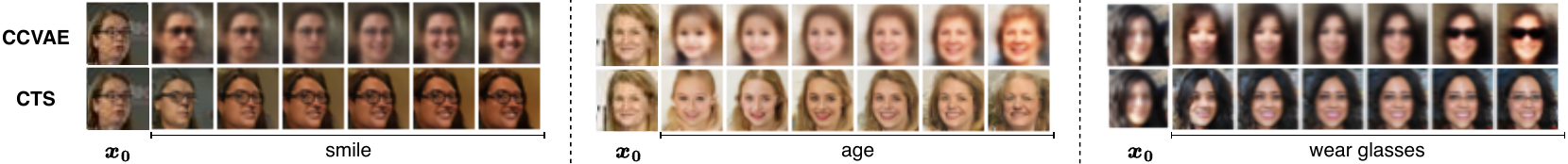}}
\vspace{-1.0em}
\caption{Explanability: a case study on FFHQ.}
\label{fig:explanability}
\vspace{-0.5em}
\end{figure*}

\subsection{Explainability}
\label{sec:expt:explanability}

Given that images are intuitively more understandable to humans than time series, we undertake a case study using a subset of FFHQ with eight external conditions to showcase the explainability of \textsf{CTS} regarding Data Selection and Condition Mapping. 

\paragraph{Explainability of Data Selection}
The top row of Figure \ref{fig:case_ffhq} displays eight clusters chosen by \textsf{DCS}. 
In each cluster, the center is marked by an orange line,
the user-input condition vector $\bm{c_0}$ is denoted by a red dashed line, and the external condition vectors $\bm{c}$'s of the images are shown with blue lines. 
The first four clusters, where the centers are nearest to $\bm{c_0}$, show a close alignment between the red dashed and orange lines.
In contrast, the last four with centers furthest from $\bm{c_0}$ display distinct trends, highlighting the capability of \textsf{DCS} to select clusters with diverse conditions.

The bottom row of Figure \ref{fig:case_ffhq} visualizes the eight most similar images from each cluster, demonstrating the efficacy of \textsf{NNS}. 
These images, especially from the initial four, closely mirror the input image in conditions like gender, eyeglasses, and facial expression. 
In contrast, images from the last four show differences in at least one condition relative to the input $\bm{c_0}$, underscoring \textsf{NNS}'s proficiency in identifying similar images with varied conditions.

\paragraph{Explainability of Condition Mapping}
We then explore the explainability of Condition Mapping using \textsf{Decision Tree}, visualized in Figure \ref{fig:dt}.
The top left section highlights the importance of external conditions such as \textit{Age}, \textit{Black Hair}, \textit{Blonde Hair}, \textit{Bald}, and \textit{Smile}, marking them as notably significant.
Each tree node has a vertical dashed line, where the split point is marked by a black wedge. 
The tree starts by splitting on \textit{Gender}, a binary condition adept at segmenting the latent space. 
As we traverse down, the splits become more refined and primarily focus on these key conditions, underscoring their pivotal role in shaping latent feature representation.
This structure offers a clear, rule-based interpretation of how latent features relate to external conditions.

When users enter a new external condition vector, $\bm{c_0^\prime}$ (shown in the top right of Figure \ref{fig:dt}), the Condition Mapping module traverses the tree based on $\bm{c_0^\prime}$, until to a leaf node. 
This terminal node predicts the latent vector, $\bm{\mu_0^\prime}$, granting users a clear view of how external conditions and latent features interrelate. 

\paragraph{Latent Traversals}
Finally, we visualize user control over image generation by altering external conditions. We spotlight the results from \textsf{CTS} and the runner-up, \textsf{CCVAE}, for brevity. 
Figure \ref{fig:case_walk} showcases the generated images when altering three specific conditions: \emph{Smile}, \emph{Age}, and \emph{Wear Glasses}.
The superior controllability of \textsf{CTS} is apparent, as it adeptly adjusts each condition independently while maintaining the integrity of others. 
For example, when adding a smile to a sideways-looking female, \textsf{CTS} retains the original image's orientation, while \textsf{CCVAE} changes unrelated conditions, such as removing glasses, contradicting the user's intent.
This may be due to \textsf{CCVAE}'s training across broader images, including those with dark sunglasses, in contrast to \textsf{CTS}'s Data Selection that focuses on images with similar characteristics.

%% file: tables/control_ts.tex
\begin{table}[ht]
\centering
\captionsetup{skip=0.5em}
\setlength\tabcolsep{3pt}
\caption{Controllability results on time series datasets.}
\label{tab:control-time-series}%
\resizebox{\columnwidth}{!}{%
\begin{tabular}{ccccccccccccc} \toprule
  \multirow{2}[6]{*}{\textbf{Settings}} & \multicolumn{6}{c}{\textbf{Interpolation}} & \multicolumn{6}{c}{\textbf{Extrapolation}} \\
  \cmidrule(lr){2-7} \cmidrule(lr){8-13} & \multicolumn{2}{c}{\textbf{Air-GZ}} & \multicolumn{2}{c}{\textbf{Air-SZ}} & \multicolumn{2}{c}{\textbf{Boiler}} & \multicolumn{2}{c}{\textbf{Air-GZ}} & \multicolumn{2}{c}{\textbf{Air-SZ}} & \multicolumn{2}{c}{\textbf{Boiler}} \\
  \cmidrule(lr){2-3}\cmidrule(lr){4-5}\cmidrule(lr){6-7}\cmidrule(lr){8-9}\cmidrule(lr){10-11} \cmidrule(lr){12-13} & \textbf{ACC$\uparrow$} & \textbf{W-F1$\uparrow$} & \textbf{ACC$\uparrow$} & \textbf{W-F1$\uparrow$} & \textbf{ACC$\uparrow$} & \textbf{W-F1$\uparrow$} & \textbf{ACC$\uparrow$} & \textbf{AUC$\uparrow$} & \textbf{ACC$\uparrow$}  & \textbf{AUC$\uparrow$} & \textbf{ACC$\uparrow$} & \textbf{AUC$\uparrow$}  \\ 
  \midrule
  \textsf{Validation} & 0.690 & 0.682 & 0.674 & 0.681 & 0.688 & 0.716 & 0.824 & 0.651 & 0.828 & 0.781 & 0.733 & 0.652 \\
  \cmidrule(lr){1-13}
  \textsf{Rand-LR} & 0.328 & 0.310 & 0.317 & 0.263 & 0.396 & 0.265 & 0.609 & 0.586 & 0.641 & 0.625 & 0.578 & 0.477 \\
    \textsf{Rand-RF} & 0.414 & 0.360 & 0.367 & 0.314 & 0.479 & 0.394 & 0.859 & 0.684 & 0.844 & 0.808 & 0.594 & 0.549 \\
    \textsf{Rand-DT} & 0.431 & 0.415 & 0.450 & 0.407 & 0.479 & 0.423 & 0.875 & 0.618 & 0.922 & 0.813 & \textbf{0.797} & 0.604 \\
    \textsf{DCS-LR} & 0.345 & 0.283 & 0.317 & 0.283 & 0.417 & 0.300 & 0.813 & 0.632 & 0.688 & 0.696 & 0.547 & 0.577 \\
    \textsf{DCS-RF} & 0.414 & 0.397 & 0.383 & 0.358 & 0.479 & 0.409 & \textbf{0.906} & 0.701 & 0.891 & \textbf{0.858} & 0.578 & 0.573 \\
    \textsf{CTS}   & \textbf{0.500} & \textbf{0.483} & \textbf{0.500} & \textbf{0.465} & \textbf{0.542} & \textbf{0.457} & 0.891 & \textbf{0.707} & \textbf{0.938} & 0.817 & 0.688 & \textbf{0.702} \\
    \textsf{CTS}$_{\textsf{VQ}}$ & 0.276 & 0.213 & 0.250 & 0.244 & 0.417 & 0.245 & 0.828 & 0.644 & 0.828 & 0.750 & 0.641 & 0.629 \\
  \bottomrule
\end{tabular}%
}
\end{table}

%% file: tables/ablation_control.tex

\begin{table}[ht]
\centering
\captionsetup{skip=0.5em}
\setlength\tabcolsep{3pt}
\caption{Ablation study of controllability.}
\label{tab:ablation-control}%
\resizebox{\columnwidth}{!}{%
\begin{tabular}{ccccccccccccc} \toprule
  \multirow{2}[6]{*}{\textbf{Methods}} & \multicolumn{6}{c}{\textbf{Interpolation}} & \multicolumn{6}{c}{\textbf{Extrapolation}} \\
  \cmidrule(lr){2-7} \cmidrule(lr){8-13} & \multicolumn{2}{c}{\textbf{Air-GZ}} & \multicolumn{2}{c}{\textbf{Air-SZ}} & \multicolumn{2}{c}{\textbf{Boiler}} & \multicolumn{2}{c}{\textbf{Air-GZ}} & \multicolumn{2}{c}{\textbf{Air-SZ}} & \multicolumn{2}{c}{\textbf{Boiler}} \\
  \cmidrule(lr){2-3}\cmidrule(lr){4-5}\cmidrule(lr){6-7}\cmidrule(lr){8-9}\cmidrule(lr){10-11} \cmidrule(lr){12-13} & \textbf{ACC$\uparrow$} & \textbf{W-F1$\uparrow$} & \textbf{ACC$\uparrow$} & \textbf{W-F1$\uparrow$} & \textbf{ACC$\uparrow$} & \textbf{W-F1$\uparrow$} & \textbf{ACC$\uparrow$} & \textbf{AUC$\uparrow$} & \textbf{ACC$\uparrow$}  & \textbf{AUC$\uparrow$} & \textbf{ACC$\uparrow$} & \textbf{AUC$\uparrow$}  \\ 
  \midrule
    \textsf{CTS}   & \textbf{0.500} & \textbf{0.483} & \textbf{0.500} & \textbf{0.465} & \textbf{0.542} & \textbf{0.457} & \textbf{0.891} & \textbf{0.707} & \textbf{0.938} & 0.817 & \textbf{0.688} & \textbf{0.702} \\
    \textsf{CTS$-$NNS} & 0.345 & 0.343 & 0.433 & 0.433 & 0.458 & 0.381 & 0.844 & 0.615 & 0.844 & 0.817 & 0.625 & 0.604 \\
    \textsf{CTS$-$DCS} & 0.414 & 0.353 & 0.450 & 0.456 & 0.458 & 0.375 & 0.781 & 0.586 & 0.813 & \textbf{0.829} & 0.656 & 0.651 \\
    \textsf{CTS$-$NNS$-$DCS} & 0.345 & 0.305 & 0.417 & 0.410 & 0.354 & 0.278 & 0.703 & 0.592 & 0.813 & 0.796 & 0.594 & 0.579 \\
  \bottomrule
\end{tabular}%
}
\end{table}

%% file: tables/control_image.tex
\begin{table}[ht]
\centering
\captionsetup{skip=0.5em}
\small
\setlength\tabcolsep{3pt}
\caption{Controllability results on image datasets.}
\label{tab:control-image}%
\resizebox{\columnwidth}{!}{%
  \begin{tabular}{ccccccccccc}
  \toprule
  \multirow{1.5}[4]{*}{\textbf{Methods}} & \multicolumn{3}{c}{\textbf{CelebA}} & \multicolumn{3}{c}{\textbf{FFHQ}} & \multicolumn{2}{c}{\textbf{Teapots}} \\
  \cmidrule(lr){2-4}\cmidrule(lr){5-7}\cmidrule(lr){8-9}   
  & \textbf{$I_+\uparrow$} & \textbf{$I_-\downarrow$} & \textbf{MD$\downarrow$} & \textbf{$I_+\uparrow$} & \textbf{$I_-\downarrow$} & \textbf{MD$\downarrow$} & \textbf{PSNR$\uparrow$} & \textbf{SSIM$\uparrow$} \\
  \midrule
  \textsf{CTS} & \textbf{0.399} & \textbf{0.143} & 0.072  & \textbf{0.403} & \textbf{0.098} & 0.084 & \textbf{25.298} & \textbf{0.857}  \\
  \textsf{CCVAE} & 0.392  & 0.146  & \textbf{0.068} & 0.399  & 0.108  & \textbf{0.063} & 20.133  & 0.746  \\
  \textsf{Guided-VAE} & 0.374  & 0.151  & 0.091  & 0.368  & 0.121  & 0.083  & 18.143  & 0.731   \\
  \textsf{IDVAE} & 0.322  & 0.164  & 0.128  & 0.326  & 0.148  & 0.070  & 14.513  & 0.562   \\
  \bottomrule
  \end{tabular}%
}
\end{table}

%% file: tables/generalization_test_image.tex
\begin{table*}[t]
\centering
\captionsetup{skip=0.5em}
\small
\setlength\tabcolsep{3pt}
\caption{Generalization test on image datasets.}
\label{tab:generalization-image}%
\resizebox{\textwidth}{!}{%
\begin{tabular}{ccccccccccccccccccccc} \toprule
  \multirow{2}[6]{*}{\textbf{\textsf{CTS} with}} & \multicolumn{7}{c}{\textbf{CelebA}} & \multicolumn{7}{c}{\textbf{FFHQ}} & \multicolumn{6}{c}{\textbf{Teapots}} \\ 
  \cmidrule(lr){2-8}\cmidrule(lr){9-15}\cmidrule(lr){16-21} & \multicolumn{2}{c}{\textbf{Gen. Fid.}} & \multicolumn{2}{c}{\textbf{Percep. Sim.}} & \multicolumn{3}{c}{\textbf{Control.}} & \multicolumn{2}{c}{\textbf{Gen. Fid.}} & \multicolumn{2}{c}{\textbf{Percep. Sim.}} & \multicolumn{3}{c}{\textbf{Control.}} & \multicolumn{2}{c}{\textbf{Gen. Fid.}} & \multicolumn{2}{c}{\textbf{Percep. Sim.}} & \multicolumn{2}{c}{\textbf{Control.}} \\
  \cmidrule(lr){2-3} \cmidrule(lr){4-5} \cmidrule(lr){6-8} \cmidrule(lr){9-10} \cmidrule(lr){11-12} \cmidrule(lr){13-15} \cmidrule(lr){16-17} \cmidrule(lr){18-19} \cmidrule(lr){20-21} 
  & \textbf{PSNR$\uparrow$} & \textbf{SSIM$\uparrow$} & \textbf{FID$\downarrow$} & \textbf{LPIPS$\downarrow$} & \textbf{$I_+\uparrow$} & \textbf{$I_-\downarrow$} & \textbf{MD$\downarrow$} & \textbf{PSNR$\uparrow$} & \textbf{SSIM$\uparrow$} & \textbf{FID$\downarrow$} & \textbf{LPIPS$\downarrow$} & \textbf{$I_+\uparrow$} & \textbf{$I_-\downarrow$} & \textbf{MD$\downarrow$} & \textbf{PSNR$\uparrow$} & \textbf{SSIM$\uparrow$} & \textbf{FID$\downarrow$} & \textbf{LPIPS$\downarrow$} & \textbf{PSNR$\uparrow$} & \textbf{SSIM$\uparrow$} \\
  \midrule
  \textsf{DC-VAE} & \textbf{18.199} & \textbf{0.662} & 55.412  & \textbf{0.027} & \textbf{0.399} & \textbf{0.143} & \textbf{0.072} & \textbf{18.439} & \textbf{0.676} & 56.884  & \textbf{0.018} & \textbf{0.403} & 0.098  & 0.084  & \textbf{29.040} & \textbf{0.912} & \textbf{31.413} & \textbf{0.005} & \textbf{25.298} & \textbf{0.857} \\
  \textsf{Soft-IntroVAE} & 17.570  & 0.660  & \textbf{49.970} & 0.029  & 0.368  & 0.155  & 0.084 & 17.050  & 0.601  & \textbf{56.711} & 0.047  & 0.374  & \textbf{0.094} & \textbf{0.081} & 26.145  & 0.869  & 78.915  & 0.012  & 20.145  & 0.728  \\
  \textsf{VQ-VAE} & 15.920  & 0.571  & 73.470  & 0.070  & 0.344  & 0.157  & 0.088 & 16.128  & 0.606  & 80.145  & 0.103  & 0.341  & 0.114  & 0.091 & 23.791  & 0.854  & 99.145  & 0.029  & 19.421  & 0.701  \\
  \bottomrule
\end{tabular}}
\end{table*}

%% file: 09_conclusions.tex
\section{Conclusions}
\label{sec:conclusion}

In this work, we explore the emerging field of CTSG and introduce \textsf{CTS}, a trailblazing VAE-agnostic framework tailored for this purpose.
\textsf{CTS} distinctly decouples the mapping function from the conventional VAE training, incorporating transparent Data Selection and Condition Mapping modules. This design offers users refined control and deepens their comprehension of the complex dynamics between latent features and external conditions.
Through extensive experiments, we demonstrate the exceptional prowess of \textsf{CTS} in generating high-quality, controllable time series. Additionally, its application in the image domain showcases its versatility and superior explainability. 
Crucially, \textsf{CTS} sheds light on the intricate interplay between latent features and external conditions, enhancing interpretability in generative modeling and providing profound insights to researchers and practitioners across various fields, thereby addressing the challenges of data scarcity.